\documentclass{article}

\usepackage[nonatbib, final]{neurips_2025}
\usepackage[square,numbers,sort]{natbib}
\usepackage[utf8]{inputenc} %
\usepackage[T1]{fontenc}    %
\usepackage[colorlinks]{hyperref}       %
\usepackage{url}            %
\usepackage{url}            %
\usepackage{float}         %
\usepackage{booktabs}       %
\usepackage{amsfonts}       %
\usepackage{nicefrac}       %
\usepackage{microtype}      %
\usepackage{xcolor}         %
\usepackage{multirow}
\usepackage{amssymb}        %
\usepackage[table]{xcolor}
\usepackage[dvipsnames]{xcolor}
\usepackage{tabularx}
\usepackage{graphicx}
\usepackage{caption}
\usepackage{subcaption}
\usepackage{makecell} 
\usepackage{fp}
\usepackage{xspace}
\usepackage{amsmath}
\usepackage{tcolorbox}
\usepackage{soul}
\usepackage{wrapfig}
\usepackage{pifont}
\usepackage{enumitem}
\usepackage{cleveref}
\usepackage{xcolor} 
\usepackage{graphicx}

\newcommand{\ie}{\textit{i.e.\xspace}}
\newcommand{\eg}{\textit{e.g.\xspace}}

\newcommand{\ourdataset}{ConViS-Bench\xspace}
\newcommand{\ourtask}{Concept-based Video Similarity estimation\xspace}
\newcommand{\ourtaskshort}{ConViS\xspace}
\newcommand{\numvideopairs}{610~}
\usepackage{pifont,xcolor}
\newcommand{\cmark}{\textcolor{green!60!black}{\ding{51}}}%
\newcommand{\xmark}{\textcolor{red!60!black}{\ding{55}}}%

\definecolor{ultraviolet}{HTML}{8365BA}

\usepackage{fp,xcolor,colortbl}
\usepackage{ifthen}
\usepackage{fp}
\usepackage{colortbl}
\definecolor{CustomRed}{rgb}{0.99, 0.72, 0.72}
\definecolor{CustomGreen}{rgb}{0.72, 0.94, 0.72}
\definecolor{DatasetPurple}{RGB}{225, 213, 231}

\newcommand{\ccc}[1]{%
  \ifthenelse{\equal{#1}{-}}{%
    \cellcolor{CustomRed!150}--%
  }{%
    \FPeval\res{clip(#1)}%
    \ifdim\res pt< -20pt
      \cellcolor{CustomRed!150}%
    \else\ifdim\res pt< -10pt
      \cellcolor{CustomRed!100}%
    \else\ifdim\res pt< 0pt
      \cellcolor{CustomRed!50}%
    \else\ifdim\res pt> 20pt
      \cellcolor{CustomGreen!150}%
    \else\ifdim\res pt> 10pt
      \cellcolor{CustomGreen!100}%
    \else\ifdim\res pt> 0pt
      \cellcolor{CustomGreen!50}%
    \else
      \cellcolor{white}%
    \fi\fi\fi\fi\fi\fi
    #1%
  }%
}

\definecolor{CustomRed2}{rgb}{1.0, 0.85, 0.85}  %
\definecolor{CustomRed1}{rgb}{1.0, 0.75, 0.75}  %
\definecolor{CustomOrange}{rgb}{1.0, 0.9, 0.7}  %
\definecolor{CustomYellow}{rgb}{1.0, 1.0, 0.8}  %
\definecolor{CustomGreen}{rgb}{0.85, 1.0, 0.85} %
\definecolor{CustomTab4}{rgb}{0.902, 0.957, 0.945}
\definecolor{CustomGreen2}{rgb}{0.624, 0.863, 0.624}

\newcommand{\colorsim}[1]{%
  \FPeval\res{clip(#1)}%
  \ifdim\res pt<20pt
    \cellcolor{CustomRed1}#1%
  \else\ifdim\res pt<30pt
    \cellcolor{CustomRed2}#1%
  \else\ifdim\res pt<40pt
    \cellcolor{CustomOrange}#1%
  \else\ifdim\res pt<50pt
    \cellcolor{CustomYellow}#1%
  \else
    \cellcolor{CustomGreen}#1%
  \fi\fi\fi\fi
}

\definecolor{promptcolor}{HTML}{D1D0F2}
\definecolor{promptcolorheader}{HTML}{9694e1}
\newcommand{\promptbox}[2]{
    \begin{tcolorbox}[
        top=0.3em,bottom=0.3em,left=0.5em,right=0.5em,
        toptitle=0.3em,bottomtitle=0.2em,boxsep=0pt,
        colframe=promptcolorheader!80,colback=promptcolor!30,boxrule=0.5pt,
        title={\footnotesize #1}
    ]
        \footnotesize
        #2
    \end{tcolorbox}
}

\newcommand{\inlineColorbox}[2]{\begingroup\setlength{\fboxsep}{1pt}\colorbox{#1}{\hspace*{2pt}\vphantom{Ay}#2\hspace*{2pt}}\endgroup}

\title{ConViS-Bench: Estimating Video Similarity Through Semantic Concepts}

\author{%
  Benedetta Liberatori\textsuperscript{1,}\thanks{Correspondence to: \texttt{benedetta.liberatori@unitn.it}.}\quad
  Alessandro Conti\textsuperscript{1}\quad
  Lorenzo Vaquero\textsuperscript{2}\quad \vspace{.1em}\\
  \textbf{Yiming Wang\textsuperscript{2}}\quad
  \textbf{Elisa Ricci\textsuperscript{1,2}} \quad
  \textbf{Paolo Rota\textsuperscript{1}}\quad \vspace{.1em}\\
  \textsuperscript{1}University of Trento \vspace{.1em}\\ \textsuperscript{2}Fondazione Bruno Kessler (FBK)
}

\begin{document}

\maketitle
\vspace{-1cm}
\begin{figure}[h!]
    \centering
    \includegraphics[width=0.80\linewidth]{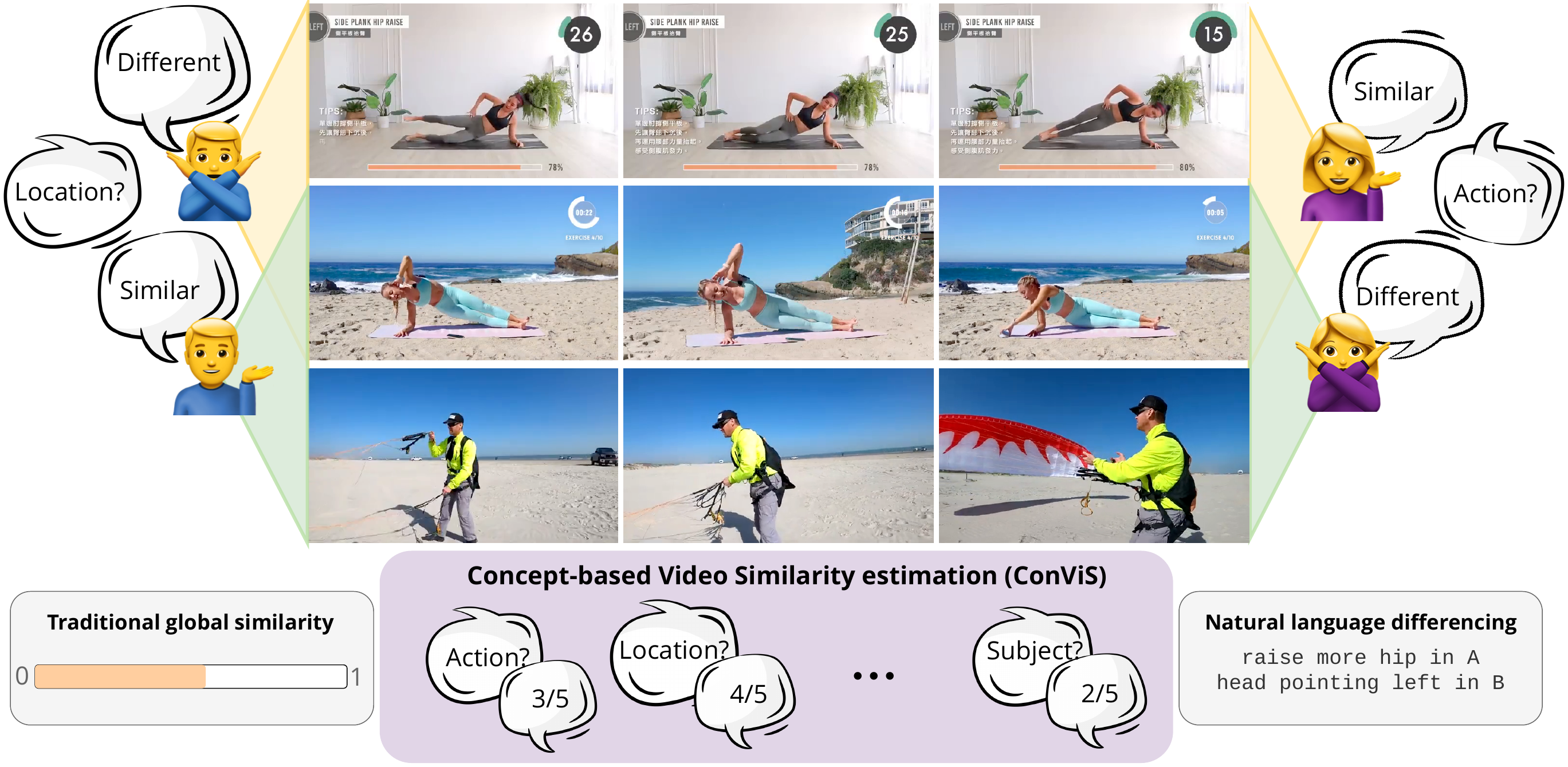}
    \caption{\textbf{\ourtask\ (\ourtaskshort).} We introduce \ourtaskshort, a task that quantifies video similarity along specific semantic concepts (\eg, location), and \inlineColorbox{DatasetPurple}{\ourdataset{}}, a dataset of video pairs annotated with concept-level similarity scores (1-to-5) and free-form descriptions of similarities and differences. This bridges the gap between prior work focused solely on \inlineColorbox{gray!20}{global similarity}~\cite{KordopatisZilos2022,KordopatisZilos2023,VVSJo2024} or \inlineColorbox{gray!20}{differences in natural language}~\cite{nagarajan2024step, burgess2025video}.}
    \label{fig:teaser}
\end{figure}

\begin{abstract}
What does it mean for two videos to be similar? Videos may appear similar when judged by the actions they depict, yet entirely different if evaluated based on the locations where they were filmed. While humans naturally compare videos by taking different aspects into account, this ability has not been thoroughly studied and presents a challenge for models that often depend on broad global similarity scores.
Large Multimodal Models (LMMs) with video understanding capabilities open new opportunities for leveraging natural language in comparative video tasks.
We introduce \ourtask (\ourtaskshort), a novel task that compares pairs of videos by computing interpretable similarity scores across a predefined set of key semantic concepts. 
\ourtaskshort allows for human-like reasoning about video similarity and enables new applications such as concept-conditioned video retrieval. 
To support this task, we also introduce \ourdataset, a new benchmark comprising carefully annotated video pairs spanning multiple domains. 
Each pair comes with concept-level similarity scores and textual descriptions of both differences and similarities.
Additionally, we benchmark several state-of-the-art models on \ourtaskshort, providing insights into their alignment with human judgments. 
Our results reveal significant performance differences on \ourtaskshort, indicating that some concepts present greater challenges for estimating video similarity. We believe that \ourdataset will serve as a valuable resource for advancing research in language-driven video understanding. Code and dataset are available at \url{https://github.com/benedettaliberatori/convisbench}.

\end{abstract}
\section{Introduction}\label{sec:intro}

Humans can effortlessly compare pairs of videos, rapidly identifying similarities and differences by attending to a range of semantic aspects, such as the depicted activity, the people involved, or the environment. This intuitive comparative ability relies on a rich understanding of events that unfold across space and time.
Cognitive science research confirms this: humans naturally perceive, encode, and retrieve events along key semantic concepts, selectively attending to particular attributes of experience~\cite{sastre2023binds,Wahlheim2025-xm}. As a result, perceived similarity between two videos is not a fixed quantity, but it depends on which concepts are being prioritized. For example, two videos may appear highly similar in terms of the activity being performed, but diverge substantially in the setting or the agents involved (see Fig.~\ref{fig:teaser}, top). This observation motivates a shift from holistic similarity to a more structured, concept-aware notion of video comparison.

Computational approaches to video similarity have traditionally focused on global similarity metrics, typically learned by comparing spatio-temporal embeddings~\cite{KordopatisZilos2022,KordopatisZilos2023,VVSJo2024}. 
The emergence of Large Multimodal Models (LMMs)~\cite{li2025llavaonevision,lillava,ye2025mplugowl,zhu2025internvl3,bai2025qwen2} with video understanding capabilities has opened new possibilities for using natural language to describe and reason about differences between videos. Prior work has explored this by generating natural language descriptions of video differences, either through domain-specific cooking concepts~\cite{nagarajan2024step} or fine-grained, action-specific skill differences~\cite{burgess2025video}. 
However, these approaches remain limited to narrow domains and are purely descriptive, lacking \textit{structured, quantitative assessments} of similarity across semantic concepts.
As a result, comparative video understanding \textit{via language} remains in its early stages, with existing benchmarks failing to capture the broad semantic diversity present in real-world scenarios.

To address this gap, we introduce a new task, \emph{\ourtask} (\ourtaskshort). 
Inspired by human cognition and grounded in semantic structure, \ourtaskshort aims to quantify how similar two videos are on specific concepts, \eg, the activity, the location, or the order of actions (see Fig.~\ref{fig:teaser}, bottom).
\ourtaskshort enables concept-specific video understanding, supporting applications like targeted video retrieval (\eg, same activity with different subjects), anomaly detection based on particular factors (\eg, unusual object presence or action sequence), and fine-grained model evaluation by isolating the conceptual sources of failure (\eg, confusing similar-looking scenes with different actions). 
Building on the definition of \ourtaskshort, we introduce a novel benchmark, \emph{\ourdataset}, to support model evaluation and foster further research. \ourdataset consists of video pairs spanning a broad range of domains, each annotated by multiple human evaluators with similarity scores conditioned on multiple semantic concepts and accompanied by textual descriptions.

Alongside introducing a novel dataset associated with the newly proposed task, we extensively benchmark several recent LMMs to assess their ability in predicting concept-based video similarities. 
Our analysis of their relevance to human judgment reveals significant performance differences across various LMMs on \ourtaskshort, highlighting that certain concepts are more challenging for models to judge in terms of video similarity. 
For instance, while some models can reliably identify visual similarities, they consistently struggle with more abstract notions such as the temporal structure of events, an issue also noted in prior work~\cite{bagad2023test,lei-etal-2023-revealing}.
Lastly, we demonstrate the utility of concept-aware similarity in downstream tasks such as concept-conditioned video-to-video retrieval, showing how \ourtaskshort can enable nuanced and interpretable video analysis.

Overall, our contributions are threefold:
\begin{itemize}[leftmargin=3em]
    \item We introduce the \ourtaskshort task, a new formulation of video similarity that moves beyond traditional global scoring and computes interpretable similarity scores across semantic concepts.
    \item We release \ourdataset, a new benchmark dataset with human-annotated similarity judgments across multiple semantic concepts and diverse video domains.
    \item We conduct an extensive evaluation of state-of-the-art (video- and image-based) models on \ourdataset, analyzing their current strengths and limitations in concept-aware video comparison.
\end{itemize}

\section{Related Work}
\vspace{-0.2cm}
Our work is related to previous research on comparing pairs of images and videos using natural language. We also discuss previous studies aimed at assessing the capabilities of LMMs in several video understanding tasks.

\textbf{Visual Differences in Images.} 
Image change captioning refers to the task of describing the differences between two images by generating a sentence caption~\cite{jhamtani-berg-kirkpatrick-2018-learning,hosseinzadeh2021image,kim2021agnostic,yao2022image}.
Recent advancements in LMMs have enabled the joint analysis and comparison of multiple images~\cite{jiang2024mantis,li2025llavaonevision,chen2024expanding,bai2025qwen2,ye2025mplugowl}. 
However, benchmarking studies indicate that model performance on multi-image understanding tasks remains relatively low, especially in tasks involving spatial understanding~\cite{meng2025mmiu} and fine-grained visual distinctions~\cite{tong2024eyes}.
Closely related to our work is~\cite{achille2024interpretable}, which defines a notion of \emph{conceptual similarity} between pairs of images. This concept captures high-level relations, even between images that do not share visually similar elements. Our work tackles a related but more ambitious challenge, as analyzing videos is inherently harder than comparing images.

\textbf{Visual Differences in Videos.}
Traditional approaches comparing videos typically compute a single, global similarity score~\cite{KordopatisZilos2022,KordopatisZilos2023,VVSJo2024}.
While these methods are practical, the resulting scores are challenging to interpret. 
To address this, recent research has shifted towards natural language-based video differencing~\cite{burgess2025video,nagarajan2024step}.
\citet{burgess2025video} introduces the task of Video Action Differencing and the associated VidDiffBench dataset, aimed at identifying subtle differences in how individuals perform the same action. Their approach focuses on textual descriptions of fine-grained skill variations, rather than quantifying similarity with explicit scores.
Similarly, \citet{nagarajan2024step} proposes the Difference Question Answering task through the StepDiff dataset, which targets cooking-related instructional videos.
Here, the model compares two videos of the same procedural step and optionally ranks videos based on a common reference. 
While these approaches advance research in video pair comparison, they are not tailored to support applications that necessitate concept-specific similarity quantification, such as those discussed in Sec.~\ref{sec:intro}.
Moreover, the accompanying datasets are typically limited to a few domains. In contrast, we advocate the need for datasets and methods that generalize across a broader range of video types and domains. 

Our work is also related to composed video retrieval~\cite{ventura2024covr,hummel2024egocvr}, where a query consists of a reference video plus a textual modification, and the goal is to retrieve a target video that reflects that modification.
However, our task differs in two principal ways: (i) rather than focusing on retrieving a matching video, we aim to compute explicit, quantitative similarity scores; and (ii) we support exploration along multiple conceptual dimensions, enabling comparisons not limited to a single textual modification.

\textbf{Benchmarking LMMs for videos.}
The rapid advancement of LMMs have driven a surge in the number of benchmarks used to evaluate their video understanding capabilities across spatial, temporal, and semantic dimensions. 
For instance, MMBench-Video~\cite{fang2024mmbench}, MVBench~\cite{li2024mvbench}, and VideoVista~\cite{li2024videovista} assess LMMs’ ability to reason temporally, semantically, and causally across varied video content. Similarly, Video-MME~\cite{fu2024video} and Video-MMMU~\cite{hu2025video} provide full-spectrum evaluations over a wide range of video domains, durations, and task complexities, with the latter explicitly modeling cognitive stages such as perception, comprehension, and adaptation.
Several benchmarks also probe specific aspects of video understanding. TempCompass~\cite{liu2024tempcompass} targets fine-grained temporal perception by constructing videos with identical static content but controlled temporal variations, while PerceptionTest~\cite{patraucean2023perception} examines high-level reasoning across physical and counterfactual dimensions. 
Benchmarks like LVBench~\cite{wang2024lvbench} and ActivityNet-QA~\cite{yu2019activitynet} assess long-form video understanding by testing models’ memory and narrative coherence.
These benchmarks primarily assess video understanding capabilities through multi-choice question answering. While this is effective for standardized evaluation, recent work has shown that such questions are often overly informative, leading to biased evaluations~\cite{xiao2024can,cores2024tvbench}. Furthermore, our focus here is on assessing the ability of different LMMs to capture a set of specific concepts that cognitive studies~\cite{sastre2023binds,Wahlheim2025-xm} have shown to be crucial for human understanding of videos.

\section{\ourtask}\label{sec:formalization}

Humans naturally compare videos by focusing on specific semantic aspects rather than their entire content.
For example, given the top two clips shown in Fig.~\ref{fig:teaser}, one viewer may judge them to be highly similar, as both depict the same action (\ie, side plank), while another viewer may judge them to be completely different, as the locations vary significantly (\ie, house interior \textit{vs.} beach).
Such concept-specific video comparison is crucial for applications that require the isolation of individual concepts, as discussed in Sec.~\ref{sec:intro}.

Formally, let $\mathcal{C}=\{C_1,\dots,C_K\}$ be a set of predefined concepts, each expressed in natural language (\eg, \emph{main action} or \emph{location}).
Given two videos $V_1$, $V_2$ and a concept $C_i \in \mathcal{C}$, we define the \emph{concept‑based similarity}:
\begin{equation}
  s(V_1,V_2 \mid C_i) \in \mathbb{R} 
\end{equation}
which measures how similar the two videos are to the concept $C_i$.
This formulation forces the comparison to attend only to details relevant to $C_i$, ignoring other information.

Since concepts are expressed through natural language, our approach offers two key advantages.
The first aspect is flexibility, allowing users to introduce any semantic dimension without being confined to a fixed taxonomy. The second is composability, where users can select a set of concepts and assign non-negative weights $\lambda_i$ to each concept. The individual scores can be aggregated into an overall similarity score:
\begin{equation}
  s(V_1,V_2) = \sum_{i=1}^K \lambda_i\,s(V_1,V_2 \mid C_i) \quad \text{with }\sum_{i=1}^K\lambda_i=1, \label{eq:glob_sim}
\end{equation}
which enables fine‑grained control over which aspects drive the final judgment.

Our concept‑based formulation bridges two extremes of prior research on video pair comparison.
Traditional \emph{global similarity} approaches~\cite{KordopatisZilos2022,KordopatisZilos2023} compute an unconditioned score $s(V_1,V_2)$ (\eg, via average-pooled video embeddings), but do not indicate which factors influence the similarity.
Conversely, \emph{video differencing} techniques describe qualitative differences in natural language~\cite{nagarajan2024step,burgess2025video} without yielding per‑concept quantitative scores. 
By contrast, our formulation provides both \emph{structured conditioning} on $C_i$ and \emph{interpretable quantitative similarities}, facilitating comparisons and direct evaluation of a model's alignment on user‑defined concepts.

\section{The \ourdataset Dataset}\label{sec:dataset}

We seek to advance video similarity estimation by introducing a new dataset that captures the complex, concept-dependent nature of how humans compare videos.
Existing datasets describing video pairs with language are limited in scope: some focus narrowly on a single concept, such as \textit{action}~\cite{burgess2025video}, while others are restricted to domain-specific categories, such as \textit{ingredients}~\cite{nagarajan2024step} (see Tab.~\ref{tab:dataset_comparison}).
To address this gap, we introduce \ourdataset, a benchmark for multi-concept video similarity estimation. 

\begin{figure}
    \centering
    \includegraphics[width=1\linewidth]{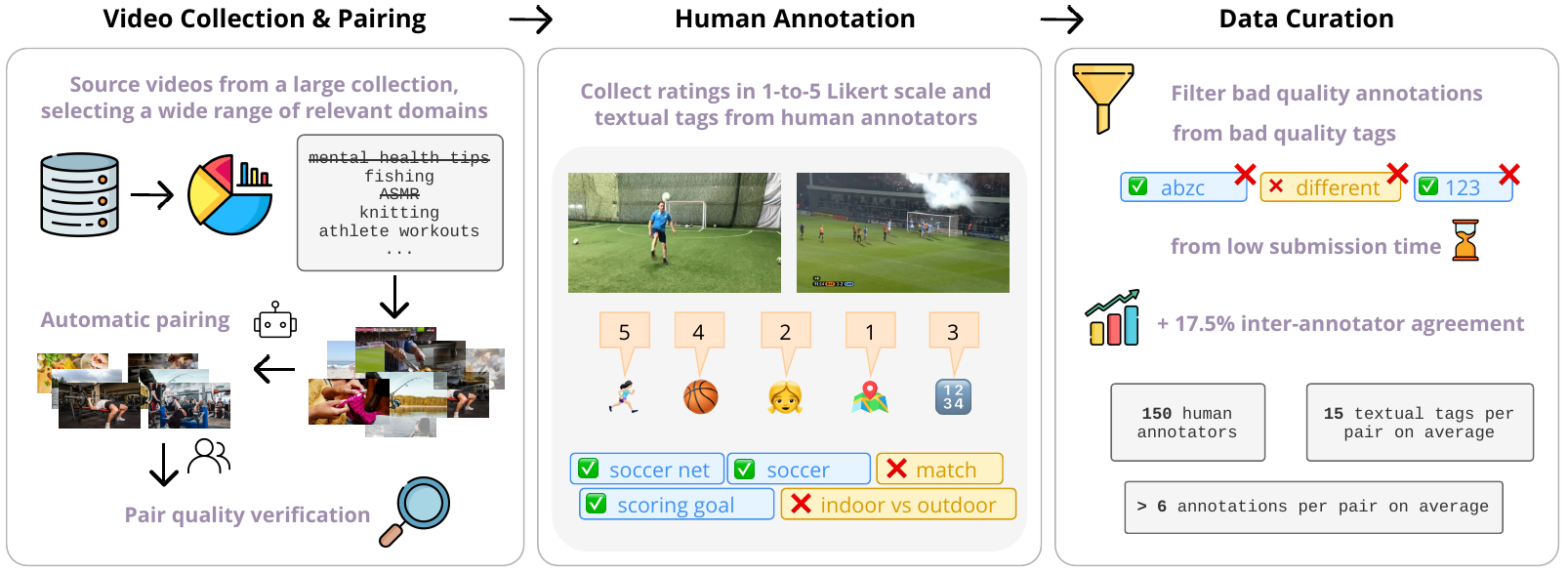}
    \caption{\textbf{The construction \ourdataset.} We
    source videos from FineVideo~\cite{FineVideo}, a large-scale collection of videos, and perform automatic trimming and pairing. We then collect human ratings on a 1-to-5 Likert scale across multiple concepts ( \raisebox{-1.0mm}{\includegraphics[height=3.6mm]{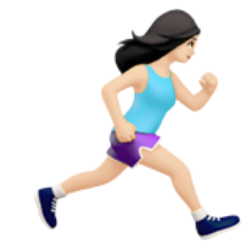}}~\textit{main~action}, \raisebox{-1.0mm}{\includegraphics[height=3.6mm]{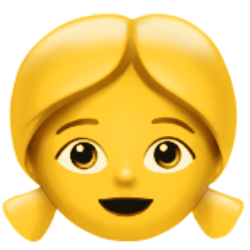}}~\textit{main~subjects}, \raisebox{-1.0mm}{\includegraphics[height=3.6mm]{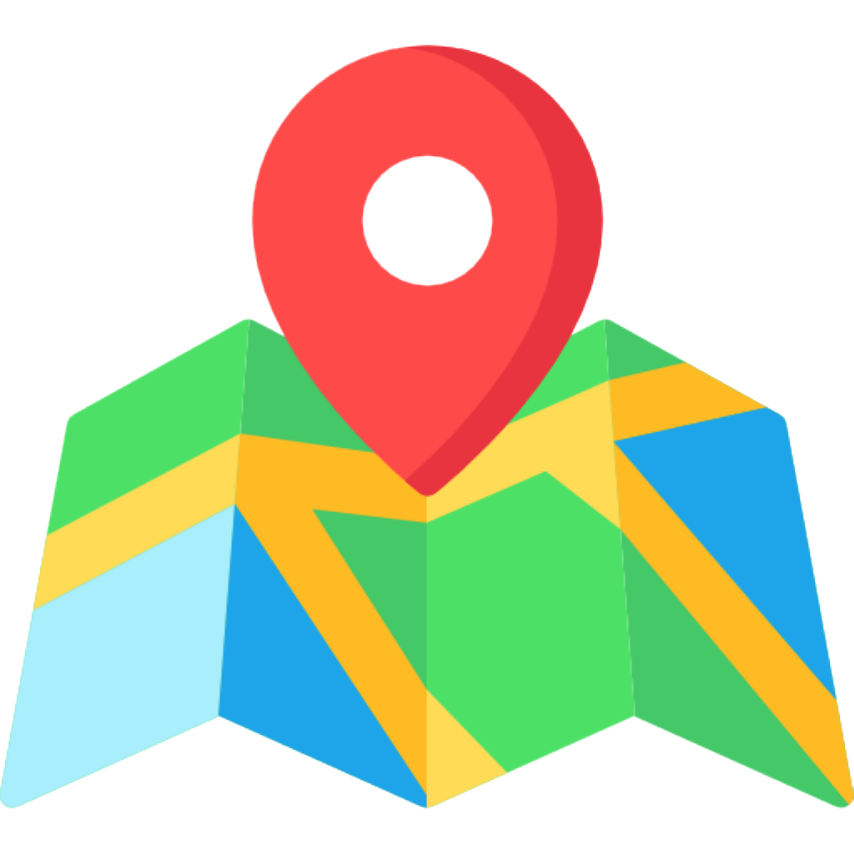}}~\textit{location}, \raisebox{-1.0mm}{\includegraphics[height=3.6mm]{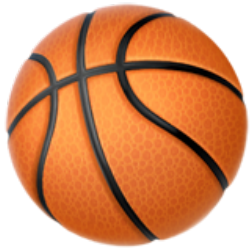}}~\textit{main~objects}, \raisebox{-1.0mm}{\includegraphics[height=3.6mm]{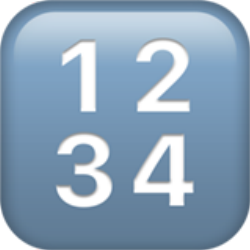}}~\textit{order~of~actions}) and free-form tags describing key similarities and differences. 
    Finally, we systematically curate the data to ensure quality. }
    \label{fig:data-collection-pipeline}
\end{figure}

We follow a structured protocol to build and curate \ourdataset (see Fig.~\ref{fig:data-collection-pipeline}).
\ourdataset contains \numvideopairs video pairs drawn from 543 videos, each annotated with human-judged similarity scores across five general-purpose concepts.
In addition to quantitative similarity scores, each pair is accompanied by free-form descriptions highlighting shared and differing elements, offering qualitative insight into human reasoning. 
It covers the broadest range of domains to date, \ie, 16 in total, significantly more than the 5 covered in~\cite{burgess2025video}, and features longer videos on average.

\begin{table}[t]
\scriptsize
\centering
\caption{\textbf{Comparing \ourdataset with prior work.} Video pairs in \ourdataset are annotated by humans with similarity scores across fine concepts and free-form text descriptions of similarities and differences. $^\dagger$StepDiff provides scores only for a subset of its domain-specific categories. }
  \begin{tabular}{@{}lccccccc@{}}
    \toprule
    Dataset        & Pairs & Domains & Focus & Scores & Text     & Split       & Avg.\ duration \\
    \midrule
    StepDiff~\cite{nagarajan2024step}   & 6,292     & 1  & 5 cooking categories   & ~~~\cmark$^\dagger$ & differences     & train/test & 12.0 s \\
    VidDiffBench~\cite{burgess2025video} & 557      & 5  & 1 concept (action)  & \xmark         & differences     & test       & 8.8 s  \\
    \ourdataset (Ours)           & 610              & 16 & 5 broad concepts  & \cmark        & similarities/differences & test       & 28.2 s \\
    \bottomrule
  \end{tabular}
    \label{tab:dataset_comparison}
\end{table}

\subsection{Collecting and annotating video pairs for \ourtaskshort}

Here, we outline the key steps involved.
Further details about the dataset statistics and the annotation process are in the Appendix. 

\textbf{Choice of Concepts.}
Our approach to computing video similarity is based on defining key semantic concepts. 
This design is motivated by research in cognitive science showing that memory representations are organized around semantic and temporal features, and that people update their understanding of events by comparing such structured elements via mechanisms like prediction error and recursive reminding~\cite{Wahlheim2025-xm,sastre2023binds}.
Guided by these findings, we identify five core concepts that reflect how people naturally interpret events: \textit{main action}, \textit{main subjects}, \textit{main objects}, \textit{location}, and \textit{order of actions}.
These constitute the set $\mathcal{C}$ as defined in Sec.~\ref{sec:formalization}, which is designed to be domain-agnostic, ensuring consistent estimation of video pair similarity across diverse domains.

\textbf{Video Collection.}
As shown in the left part of Fig.~\ref{fig:data-collection-pipeline}, \ourdataset's data collection pipeline begins by identifying an existing dataset to serve as the primary video source.
We select the FineVideo dataset~\cite{FineVideo}, a large-scale collection of YouTube videos that spans a diverse range of human activities. The dataset is densely annotated, recently compiled, and available under the Attribution (CC-BY) license. 
It includes 126 fine-grained categories organized within a two-level taxonomy, comprising 9 coarse-grained categories. 
However, not all of them are equally informative for assessing similarity in video content.
In particular, some categories, \eg, \textit{Mental Health Tips} or \textit{Poetry Reading}, exhibit limited visual diversity and temporal dynamics, as they predominantly consist of static shots of people speaking.
Since our goal is to analyze visual similarity across semantically grounded dimensions (\ie, the concepts), we restrict our focus to categories with rich, varied visual content and exhibit sufficient temporal dynamics.
Accordingly, we select a subset of relevant coarse- and fine-grained categories from FineVideo, as illustrated in Fig.~\ref{fig:data-collection-pipeline}. 
A complete list can be found in the Appendix.
As a result, we include videos from 16 diverse domains, making \ourdataset\ the most semantically and visually diverse dataset of video pairs to date, as illustrated in Tab.~\ref{tab:dataset_comparison}.

\textbf{Video Pairing.}
In FineVideo~\cite{FineVideo}, the average video length is 4.7 minutes, with each video potentially containing various shots and events.
First, to ensure compatibility, we trim the original videos using the provided timestamp annotations to isolate clips corresponding to individual events.
From this curated pool, we construct video pairs.
To promote diversity, we match videos based on either semantic similarity (guided by textual descriptions) or visual similarity. 
To this end, we compute visual embeddings for the trimmed clips and textual embeddings for their corresponding ground-truth descriptions, using DINOv2~\cite{oquab2024dinov} and Sentence-BERT~\cite{reimers2019sentencebert} as the respective encoding methods.
We then form candidate video pairs by selecting those with a high cosine similarity computed considering a single modality, either video-to-video or text-to-text, but explicitly not both.
The rationale behind this procedure is to ensure that each pair of videos shares certain common patterns while also exhibiting distinctive elements. 
Finally, we perform manual filtering and verification, obtaining \numvideopairs video pairs that exhibit varying degrees of similarity across the five considered concepts.

\textbf{Human Annotation.}
We collect human annotations for all the video pairs, as shown in the center of Fig.~\ref{fig:data-collection-pipeline}.
The goal of these annotations is to obtain reliable similarity scores for the concepts in $\mathcal{C}$, along with textual descriptions highlighting both the similarities and differences between the paired videos.
We collect annotations from a mix of volunteers and paid workers (totaling 150 annotators) recruited via Prolific, a crowdsourcing platform selected for its demonstrated quality and reliability compared to existing alternatives~\cite{douglas2023data,Peer2022-xd}.
We instruct annotators (all college-educated English speakers) to evaluate each video pair along the five predefined semantic concepts in $\mathcal{C}$.
For each concept, similarity is rated on a 5-point Likert scale~\cite{Likert1932}, with detailed scoring guidelines provided to ensure clarity and consistency across annotators. 
To further promote thoughtful engagement with the task, annotators must assign at least one \textit{similarity tag} and one \textit{difference tag} to each pair, describing specific aspects in which the videos align or differ. 

Acknowledging that our predefined concepts may not fully encompass all relevant aspects of similarity, we encourage annotators to introduce new concepts when applicable. 
Ultimately, only 5.28\% of the annotations involve a new concept, supporting our initial assumption that the predefined ones adequately capture the key aspects of similarity across videos.  
The annotation interface, statistics on inter-annotator agreement, and additional details on this core step are available in the Appendix.

\noindent\textbf{Data Curation.} As shown in the right part of Fig.~\ref{fig:data-collection-pipeline}, the collected annotations undergo a final curation process.
Specifically, we use the free-form text tags provided by annotators to assess annotation quality, removing samples containing non-English or irrelevant content, while allowing for correctable typographical errors. We also flag and manually review annotations submitted in unusually short timeframes, which may indicate rushed or low-effort responses.
As a result, we discard 7.75\% of the original annotations, resulting in a cleaner and more reliable dataset.
The final collection contains video pairs collected from 147 annotators, with an average of 6.2 annotations per pair.

\section{Benchmarking Models on \ourdataset}\label{sec:experiments}
A key contribution of this work is a comprehensive evaluation of state-of-the-art models on the proposed \ourdataset benchmark.
In this section, we report the results of our experiments. 

We first consider ten state-of-the-art LMMs and evaluate their ability to compare videos explicitly accounting for specific concepts (Sec.~\ref{sec:consim}).
Then, we analyze different approaches that assess video similarity based on a global score, 
(Sec.~\ref{sec:globalsim}). 
Finally, we assess models on the task of concept-based video retrieval (Sec.~\ref{sec:cbvr}).
All experiments were conducted using up to two NVIDIA H100 GPUs. 

\subsection{Assessing LMMs performance on \ourtask}
\label{sec:consim}
In this section, we consider state-of-the-art LMMs and evaluate their capabilities to compare videos based on specific concepts.
We select nine open-source, state-of-the-art models with strong performance in video understanding, including mPLUG-Owl3~\cite{ye2025mplugowl}, LLaVA series~\cite{li2025llavaonevision,zhang2024video}, Qwen-VL series~\cite{bai2025qwen2}, and InternVL series~\cite{zhu2025internvl3,chen2024expanding}, and the closed source model Gemini (2.0-Flash)~\cite{team2023gemini}. 
These models can be prompted with questions explicitly targeting particular aspects of a video. 
Following the approach in \cite{burgess2025video}, we provide each model with concatenated frames from both videos in a pair and ask it to \textit{``[...] output a single similarity score between 1 and 5, where 1 means completely different and 5 means perfectly similar in terms of <concept>''}.
For Gemini, which natively supports multi-video within a single prompt, we supply the complete videos. The exact prompt used and additional implementation details are described in the Appendix.
The pre-training data of Gemini is private, and we do not know if the source data used for our video pairs (\ie, FineVideo) was included. On the contrary, we know that the training set of InternVL~\cite{chen2024expanding} includes it.
Nevertheless, we still decide to include these models in the evaluation to assess their performance on the new task of \ourtask.
We use Spearman's rank correlation $\rho$ and Kendall’s $\tau$ as our evaluation metrics, following prior work evaluating alignment with human judgment~\cite{achille2024interpretable,lin2024evaluating,hu2023tifa}.
Both coefficients take values in $[-1, +1]$ (here reported $\times100$) with $-1$ denoting complete disagreement with human evaluators, and $+1$ perfect agreement.

\textbf{Can LMMs judge video similarity conditioned on specific concepts?}
Table~\ref{tab:concept_based_corr} presents the results of our concept-based evaluation. 
Unsurprisingly, larger models consistently outperform their smaller counterparts, with LLaVA-OV-7B achieving the highest correlations overall. 
Interestingly, certain concepts, such as \textit{order of actions}, are inherently more challenging for all models. 
While some models perform better on other concepts, such as \textit{main objects} or \textit{location}, with LLaVA-OV-7B scoring above $58/47\%$ in $\rho/\tau$,  these correlations with human judgment are nevertheless more difficult to capture than those related to global assessments.
Surprisingly, InternVL models, despite having seen the evaluation videos during pre-training, struggle to align with human judgments.
This indicates that pre-training with interleaved video-text data alone does not equip models to accurately estimate similarity concerning specific semantic concepts. 
Rather than task-specific skill, our results may reflect how well general-purpose LLMs are compatible with human notions of conceptual similarity.

\begin{table}[t]
\def\arraystretch{1.15}
\centering
\caption{\textbf{Alignment performance of LMMs on \ourdataset.} We report Spearman's $\rho$ and Kendall's $\tau$ correlations with human judgment ($\times100$). Top-performing scores are shown in \textbf{bold}, and second-best scores are \underline{underlined}, proprietary models in \textcolor{gray}{gray}.}
\scriptsize
\vspace{\baselineskip}
\begin{tabularx}{\textwidth}{>{\raggedright\arraybackslash}p{2.5cm}>{\centering\arraybackslash}c>{\centering\arraybackslash}c@{\hskip 20pt}>{\centering}c>{\centering\arraybackslash}c@{\hskip 20pt}c>{\centering\arraybackslash}c@{\hskip 20pt}c>{\centering\arraybackslash}c@{\hskip 20pt}c@{\hskip 5pt}c}
\toprule
 & \multicolumn{2}{c@{\hskip 20pt}}{\texttt{Main Action}} & \multicolumn{2}{c}{\hspace{-20pt}\texttt{Main Subjects}} & \multicolumn{2}{c}{\hspace{-20pt}\texttt{Main Objects}} & \multicolumn{2}{c}{\hspace{-15pt}\texttt{Location}} & \multicolumn{2}{c}{\hspace{-10pt}\texttt{Order of Actions}} \\
Model & $\rho$ & $\tau$ & $\rho$ & $\tau$ & $\rho$ & $\tau$ & $\rho$ & $\tau$ & $\rho$ & $\tau$ \\
\midrule
mPLUG-Owl3-7B~\cite{ye2025mplugowl} & 30.64 & 25.12 & 20.59 & 16.84 & 28.53 & 23.47 & 21.00 & 17.00 & 23.11 & 18.74 \\
LLaVA-OV-0.5B~\cite{li2025llavaonevision} & 1.95 & 1.54 & -5.05 & -3.91 & -4.00 & -3.20 & 5.66 & 4.59 & 1.30 & 0.90 \\
LLaVA-OV-7B~\cite{li2025llavaonevision} & \textbf{51.76} & \textbf{42.48} & \textbf{48.43} & \textbf{39.14} & \textbf{58.64} & \textbf{48.17} & \textbf{58.94} & \textbf{47.72} &  \underline{41.02} & \underline{33.31} \\
LLaVA-Video-7B~\cite{zhang2024video} & 44.17 & 36.53 & 39.81 & \underline{32.81} & \underline{45.85} & \underline{37.96} & \underline{55.96} & \underline{46.33} & \textbf{41.25} & \textbf{34.05} \\
Qwen2.5-VL-3B~\cite{bai2025qwen2} & 21.84 & 18.10 & 8.62 & 6.70 & 15.20 & 12.61 & 13.44 & 11.14 & 12.79 & 10.60 \\
Qwen2.5-VL-7B~\cite{bai2025qwen2} & 37.88 & 31.28 & 17.53 & 14.43 & 26.97 & 22.26 & 23.63 & 19.17 & 23.85 & 19.61 \\
InternVL2.5-4B~\cite{wang2024lvbench} & 13.23 & 10.13 & 16.71 & 12.89 & 14.99 & 11.71 & 13.93 & 10.71 & 9.88 & 7.54 \\
InternVL2.5-8B~\cite{chen2024expanding} & 28.70 & 22.42 & 28.60 & 22.08 & 25.06 & 19.40 & 19.64 & 15.18 & 18.15 & 14.07 \\
InternVL3-8B~\cite{zhu2025internvl3} & 40.69 & 31.31 & 36.54 & 27.80 & 42.50 & 32.88 & 45.47 & 34.98 & 32.74 & 24.88 \\
\textcolor{gray}{Gemini-2.0-Flash} & \underline{\textcolor{gray}{{44.91}}} & \underline{\textcolor{gray}{34.09}} & \underline{\textcolor{gray}{41.16}} & \textcolor{gray}{31.56} & \textcolor{gray}{38.36} & \textcolor{gray}{29.21} & \textcolor{gray}{42.12} & \textcolor{gray}{34.54} & \textcolor{gray}{32.36} & \textcolor{gray}{25.00} \\
\bottomrule
\end{tabularx}
\label{tab:concept_based_corr}
\end{table}

\begin{figure}
    \centering
\includegraphics[width=\textwidth]{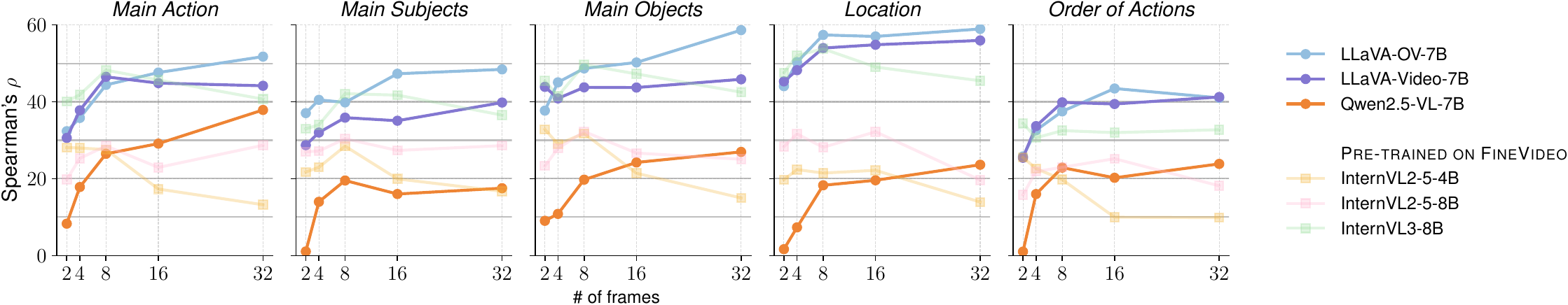} 
\caption{\textbf{Concept-based video similarity at varying number of input frames.} We report the Spearman's $\rho$ correlations with human judgement ($\times100$) for different numbers of frames per pair.}
\label{fig:var_num_frames}
\end{figure}

\textbf{Does \ourdataset challenge models on temporal understanding?}
To assess whether our benchmark poses a temporal challenge and how the temporal dimension impacts \ourtaskshort, we repeat the analysis in Tab.~\ref{tab:concept_based_corr} while varying the number of input frames per video.  
As shown in Fig.~\ref{fig:var_num_frames}, models not pre-trained on FineVideo (LLaVA-OV/LLaVA-Video/Qwen2.5-VL), consistently suffer from reduced temporal context, with performance dropping sharply as fewer frames are provided (average drops: $-16.36/10.61/21.74\%$).
In contrast, models from the InternVL family, exposed to FineVideo during pre-training, exhibit remarkably flat performance trends, suggesting limited reliance on temporal cues. 
These results indicate a likely memorization effect in these latter models, as their performance remains stable even with severely reduced temporal input. Furthermore, we observe that the \textit{main action} concept benefits the most from additional frames, with gains of $+29.57/19.40/13.56\%$ compared to $+16.45/11.37/11.09\%$ for \textit{main subjects} in Qwen2.5-VL, LLaVA-OV, and LLaVA-Video. 
These findings show that \ourdataset poses a temporal challenge, where model performance significantly depends on access to rich temporal context.

\subsection{Probing concept focus in global video representations}
\label{sec:globalsim}
In this section, we analyze which semantic concepts are implicitly prioritized by models that compute global video similarity scores.
We focus on models that extract global video representations and compare them in an embedding space, without explicit conditioning on specific concepts.
We categorize these models into three groups based on how they process and compare inputs.
\ding{172} \emph{Video-to-video} approaches like VideoMAE~\cite{NEURIPS2022_416f9cb3} and DINOv2~\cite{oquab2024dinov} use unimodal vision encoders and compute similarity directly in the embedding space.
\ding{173} \emph{Text-to-text} approaches first map videos into text format using different video captioning models~\cite{ye2025mplugowl, bai2025qwen2, chen2024expanding}, and then compare the resulting text representations in a shared embedding space using Sentence-BERT~\cite{reimers2019sentencebert}.
\ding{174} \emph{Cross-modal} approaches rely on vision-language alignment methods.
To compute the similarity between pairs of videos, we first determine the alignment score of each video with the caption generated from its counterpart using~\cite{ye2025mplugowl}. The final similarity score for a video pair is obtained by averaging these two alignment scores.
Two alignment metrics are employed: CLIPScore~\cite{hessel-etal-2021-clipscore}, which measures cosine similarity via contrastive vision-language models, and VQAScore~\cite{lin2024evaluating}, which assesses alignment through visual question-answering between captions and visual content. For CLIPScore, both CLIP~\cite{radford2021learning} and InternVideo~\cite{wang2022internvideo} serve as vision-language encoders. 
In image-based approaches, a video's representation is obtained by averaging the representations of its individual frames.

\textbf{Do global video representations implicitly align with specific concepts?}
The semantic concepts introduced in our \ourdataset benchmark enable more detailed and granular analysis and comparison across models, a distinctive feature of our proposed benchmark.
Results in Tab.~\ref{tab:global_concept_based_corr} report the outcome of our analysis.
Among the evaluated models, video-to-video approaches exhibit a stronger focus on the \textit{location} concept, with DINOv2 achieving the highest correlation with humans on this concept ($57.4/41.7$ of $\rho/\tau$). 
This indicates that spatial scene information is more effectively captured through purely visual representations.
However, these models consistently show weaker correlations in the remaining categories. 
In contrast, text-to-text models excel at capturing action-related concepts ($44.6/31.0$ and $44.3/30.9$ of $\rho/\tau$ on average for \textit{action} and \textit{main action}) but show limited ability to compare videos with respect to \textit{location} ($32.3/22.3$ of $\rho/\tau$ on average).
Notably, across all model types, the concepts of \textit{order of actions} and \textit{location} consistently yield the lowest correlations with human judgment. 
This indicates that existing models, regardless of modality, tend to overlook or inadequately model spatial context and temporal structure, two key components in understanding video semantics.
Although no single model achieves the highest alignment with human judgments across all concepts, VQAScore stands out by attaining the highest average correlation overall.

\begin{table}[t]
\def\arraystretch{1.15}
\centering
\caption{\textbf{Alignment performance computing global video representations on \ourdataset.} We report Spearman's $\rho$ and Kendall's $\tau$ correlations with human judgement ($\times100$). \textit{Video} indicates the use of a video encoder. Best-performing scores are in \textbf{bold}, and second-best scores are \underline{underlined}.}
\scriptsize
\vspace{\baselineskip}
\begin{tabularx}{\textwidth}{
>{\raggedright\arraybackslash}p{1.9cm}
>{\centering\arraybackslash}p{1.3cm}
>{\centering\arraybackslash}X
>{\centering\arraybackslash}X
>{\centering\arraybackslash}X
>{\centering\arraybackslash}X
>{\centering\arraybackslash}X
>{\centering\arraybackslash}X
}
\toprule
Model & Method & Video & \texttt{Main Action} & \texttt{Main Subjects} & \texttt{Main Objects} & \texttt{Location} & \texttt{Order of Actions} \\
 & & & $\rho/\tau$ & $\rho/\tau$ & $\rho/\tau$ & $\rho/\tau$ & $\rho/\tau$ \\
\midrule
\multicolumn{8}{l}{\textcolor{gray}{\textit{Video-to-video}}} \\
VideoMAE~\cite{NEURIPS2022_416f9cb3} & Cosine  & \checkmark & \colorsim{13.0} / 8.7 & \colorsim{23.1} / 15.6 & \colorsim{13.2} / 8.8 & \colorsim{37.8} / 26.4 & \colorsim{15.1} / 10.2 \\
DINOv2~\cite{oquab2024dinov} & Cosine  & & \colorsim{33.3} / 22.6 & \colorsim{40.9} / 28.6 & \colorsim{37.4} / 25.8 & \textbf{\colorsim{57.4}} / \textbf{41.7} & \colorsim{34.6} / 24.0 \\
\midrule
\multicolumn{8}{l}{\textcolor{gray}{\textit{Text-to-text}}} \\
mPLUG-Owl3~\cite{ye2025mplugowl} & SBERT~\cite{reimers2019sentencebert} & \checkmark & \textbf{\colorsim{52.1}} / \textbf{36.2} & \colorsim{45.5} / 31.1 & \underline{\colorsim{55.1}} / \underline{38.9} & \colorsim{28.4} / 19.6 & \textbf{\colorsim{49.9}} / \textbf{34.9} \\
InternVL2.5~\cite{chen2024expanding} & SBERT~\cite{reimers2019sentencebert} & \checkmark & \colorsim{39.4} / 27.4 & \colorsim{31.2} / 21.4 & \colorsim{45.0} / 31.1 & \colorsim{29.9} / 20.5 & \colorsim{42.0} / 29.5 \\
Qwen2.5-VL~\cite{bai2025qwen2} & SBERT~\cite{reimers2019sentencebert} & \checkmark & \colorsim{42.3} / 29.5 & \colorsim{45.4} / 31.7 & \colorsim{45.4} / 31.8 & \colorsim{38.5} / 26.7 & \colorsim{40.9} / 28.4 \\
\midrule
\multicolumn{8}{l}{\textcolor{gray}{\textit{Cross-modal}}} \\
CLIP~\cite{radford2021learning} & CLIPScore~\cite{hessel-etal-2021-clipscore} & & \colorsim{35.7} / 24.6 & \colorsim{39.0} / 27.0 & \colorsim{39.0} / 27.1 & \colorsim{36.0} / 24.9 & \colorsim{30.9} / 21.1 \\
InternVideo~\cite{wang2022internvideo} & CLIPScore~\cite{hessel-etal-2021-clipscore} & \checkmark & \colorsim{43.1} / 30.1 & \underline{\colorsim{54.8}} / \underline{38.7} & \colorsim{54.5} / 38.5 & \colorsim{40.9} / 28.4 & \colorsim{40.2} / 28.1 \\
LLaVA-OV~\cite{li2025llavaonevision} & VQAScore~\cite{lin2024evaluating} & \checkmark & \underline{\colorsim{51.1}} / \underline{36.1} & \textbf{\colorsim{55.8}} / \textbf{39.6} & \textbf{\colorsim{58.3}} / \textbf{41.4} & \underline{\colorsim{46.5}} / \underline{32.3} & \underline{\colorsim{48.1}} / \underline{33.9} \\
\bottomrule
\end{tabularx}
\label{tab:global_concept_based_corr} 
\end{table}

\subsection{Concept-conditioned video-to-video retrieval}
\label{sec:cbvr}

In this section, we evaluate whether, given an anchor video $V$ and a concept of interest $C_j\in\mathcal{C}$, models can retrieve the most conceptually similar videos from a set of related target videos $T_V=\{V_i\}_{i=1}^N$. 
Importantly, we focus on LMMs, as these methods can be conditioned on specific concepts. 
Following standard evaluation protocols~\cite{wu2023cap4video}, we report retrieval performance using recall, precision, and F\textsubscript{1} score at rank 1 (R@1, P@1, and F\textsubscript{1}@1). 
R@1 measures the percentage of queries for which a relevant video is retrieved at the top rank, while P@1 denotes the proportion of top-ranked videos that are truly relevant. The F\textsubscript{1}@1 score summarizes the trade-off between R@1 and P@1.

To conduct this analysis, we identify as anchors any video $V$ appearing in multiple annotated pairs within \ourdataset, and construct for each anchor a fixed candidate set of four videos, $T_V = \{V_i\}_{i=1}^4$. 
To ensure the reliability of these rankings and to assign similarity scores to any newly included videos, we engage three human annotators to validate or supplement the existing annotations.
We exclude the concept of \textit{order of action} from this analysis due to low annotation agreement across video sets.
While temporal ordering is clear in video pairs, aligning actions across sets introduces ambiguities, \eg, added or missing atomic actions. Empirically, annotators found it significantly more challenging to agree on consistent partial orderings across sets than within pairs.
The result is a collection of 532 concept-based partial orderings, \ie, some videos may be considered equally relevant, allowing for ties in the ranking.
For each anchor video $V$ and concept $C_j$, we compute similarity scores between $V$ and each target video $V_i \in T_V$, denoted as $s(V, V_i \mid C_j)$. These scores are then used to induce a ranking over $T_V$ for each concept.

\begin{table}[t]
\def\arraystretch{1.15}
\centering
\caption{\textbf{Concept-based retrieval performance.} We report recall, precision, and F\textsubscript{1} score at rank 1 (R@1, P@1, F\textsubscript{1}@1). The random chance baseline is in highlighted in \textcolor{gray}{gray}. }
\scriptsize
\vspace{\baselineskip}
\begin{tabularx}{\textwidth}{
>{\raggedright\arraybackslash}p{2.5cm}>{\centering\arraybackslash}X>{\centering\arraybackslash}X>{\centering\arraybackslash}X@{\hskip 20pt}X>{\centering\arraybackslash}X>{\centering\arraybackslash}X@{\hskip 20pt}X>{\centering\arraybackslash}X>{\centering\arraybackslash}X@{\hskip 20pt}X>{\centering\arraybackslash}X>{\centering\arraybackslash}X}
\toprule
Model & \multicolumn{3}{c@{\hskip 20pt}}{\texttt{Main Action}} & \multicolumn{3}{c@{\hskip 20pt}}{\texttt{Main Subjects}} & \multicolumn{3}{c@{\hskip 20pt}}{\texttt{Main Objects}} & \multicolumn{3}{c}{\texttt{Location}} \\
 & R@1 & P@1 & $F_1$@1 & R@1 & P@1 & $F_1$@1 & R@1 & P@1 & $F_1$@1 & R@1 & P@1 & $F_1$@1 \\
\midrule
mPLUG-Owl3-7B~\cite{ye2025mplugowl} & 89.8 & 44.2 & \cellcolor{CustomTab4}54.8 & 85.2 & 52.2 & \cellcolor{CustomTab4}59.0 & 91.0 & 48.9 & \cellcolor{CustomTab4}56.9 & 89.6 & 59.8 & \cellcolor{CustomTab4}66.2 \\
LLaVA-OV-0.5B~\cite{li2025llavaonevision} & 73.4 & 32.3 & \cellcolor{CustomTab4}41.2 & 69.9 & 44.1 & \cellcolor{CustomTab4}48.5 & 74.7 & 35.9 & \cellcolor{CustomTab4}45.4 & 85.3 & 51.2 & \cellcolor{CustomTab4}59.7 \\
LLaVA-OV-7B~\cite{li2025llavaonevision} & 84.2 & 54.8 & \cellcolor{CustomTab4}\textbf{61.9} & 82.0 & 66.4 & \cellcolor{CustomTab4}\textbf{68.7} & 81.8 & 58.8 & \cellcolor{CustomTab4}\underline{63.0} & 80.1 & 68.7 & \cellcolor{CustomTab4}\textbf{69.3} \\
LLaVA-Video-7B~\cite{zhang2024video} & 88.8 & 51.1 & \cellcolor{CustomTab4}\underline{60.7} & 86.6 & 63.1 & \cellcolor{CustomTab4}\underline{68.3} & 88.7 & 58.2 & \cellcolor{CustomTab4}\textbf{65.1} & 78.8 & 68.4 & \cellcolor{CustomTab4}\underline{67.5} \\
Qwen2.5-VL-3B~\cite{bai2025qwen2} & 88.0 & 42.2 & \cellcolor{CustomTab4}51.8 & 94.5 & 51.9 & \cellcolor{CustomTab4}61.6 & 96.9 & 41.7 & \cellcolor{CustomTab4}54.4 & 96.6 & 55.8 & \cellcolor{CustomTab4}66.6 \\
Qwen2.5-VL-7B~\cite{bai2025qwen2} & 63.9 & 42.5 & \cellcolor{CustomTab4}46.8 & 63.5 & 55.8 & \cellcolor{CustomTab4}54.0 & 63.5 & 44.1 & \cellcolor{CustomTab4}47.9 & 38.5 & 70.7 & \cellcolor{CustomTab4}47.1 \\
InternVL2.5-8B~\cite{chen2024expanding} & 58.8 & 47.6 & \cellcolor{CustomTab4}48.7 & 57.8 & 60.2 & \cellcolor{CustomTab4}53.6 & 58.5 & 49.9 & \cellcolor{CustomTab4}49.4 & 55.2 & 65.0 & \cellcolor{CustomTab4}53.4 \\
InternVL2.5-4B~\cite{chen2024expanding} & 60.5 & 44.6 & \cellcolor{CustomTab4}46.5 & 57.8 & 55.7 & \cellcolor{CustomTab4}50.6 & 59.9 & 47.2 & \cellcolor{CustomTab4}48.4 & 55.6 & 59.8 & \cellcolor{CustomTab4}52.0 \\
InternVL3-8B~\cite{zhu2025internvl3} & 63.0 & 58.3 & \cellcolor{CustomTab4}56.4 & 55.8 & 66.8 & \cellcolor{CustomTab4}56.0 & 57.8 & 59.7 & \cellcolor{CustomTab4}54.5 & 54.8 & 71.0 & \cellcolor{CustomTab4}56.5 \\
\midrule
\textcolor{gray}{Random Choice} & \textcolor{gray}{25.0} & \textcolor{gray}{34.4} & \cellcolor{CustomTab4}\textcolor{gray}{27.8}  &  \textcolor{gray}{25.0}  & \textcolor{gray}{44.9} & \cellcolor{CustomTab4}\textcolor{gray}{30.4} & \textcolor{gray}{25.0} & \textcolor{gray}{50.0} & \cellcolor{CustomTab4}\textcolor{gray}{31.5} &  \textcolor{gray}{25.0} &  \textcolor{gray}{36.6} & \cellcolor{CustomTab4}\textcolor{gray}{28.4} \\
Average & 74.5 & 46.4 & \cellcolor{CustomTab4}52.6 & 72.5 & 57.4 & \cellcolor{CustomTab4}58.9 & 74.7 & 49.4 & \cellcolor{CustomTab4}54.6 & 70.5 & 63.3 & \cellcolor{CustomTab4}61.1 \\
\bottomrule
\end{tabularx}
\label{tab:retireval} 
\end{table}

\vspace{+1cm}
\begin{wrapfigure}{r}{0.5\textwidth}
  \begin{center}
  \includegraphics[width=0.45\textwidth]{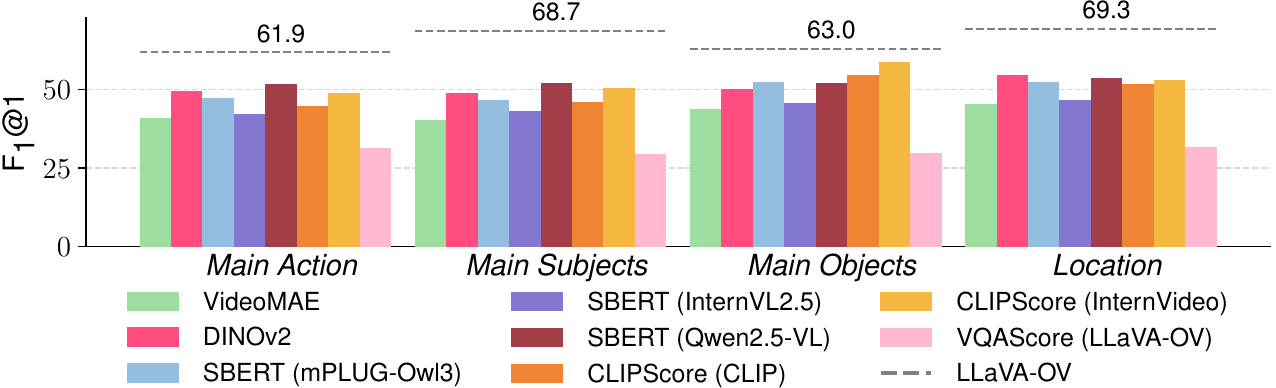}
  \end{center}
  \caption{\textbf{Retrieval performance.} F\textsubscript{1}@1 score reported for models discussed in Sec.~\ref{sec:globalsim}, against LLaVA-OV-7B~\cite{li2025llavaonevision}. }
  \label{fig:global_retrieval}
\end{wrapfigure}

\textbf{Do LMMs retrieve similarly to humans under conceptual constraints?} 
In Tab.~\ref{tab:retireval} we evaluate LMMs' performance on concept-conditioned retrieval. 
Across all models, R@1 consistently exceeds P@1, indicating that, while they often retrieve the true relevant video at rank 1, they also produce more false positives.
Performance varies across concepts: the median P@1 is higher for \textit{location} (with InternVL3-8B, LLaVA-OV-7B, LLaVA-Video-7B and Qwen2.5-VL-7B achieving P@1 $>68\%$), while \textit{main action} tends to be more challenging, with a mean P@1 of $46.4\%$.
\textit{Random choice} references a model that randomly picks one sample. Reported results take into consideration partial ordering. All evaluated models consistently outperform this random baseline in the F\textsubscript{1}@1 score, despite some having a worse precision. This indicates that the model retrieves more relevant samples overall, albeit including more irrelevant ones. 
Overall, these results indicate that while current LMMs exhibit some capabilities, concept-based video-to-video retrieval remains a limitation. 
In Fig.~\ref{fig:global_retrieval}, we compare the retrieval performance of the best-performing LMM with that of the models described in Sec.~\ref{sec:globalsim}, which are not explicitly conditioned on the concept.
While these approaches consistently underperform, the gap is smaller for \textit{main objects}, where CLIPScore achieves comparable performance, highlighting its bias toward object-level representations. 
This further highlights that tailored, concept-aware methods are essential to complement general-purpose ones for accurate specialized retrieval.

\vspace{-5pt}

\section{Conclusions}\label{sec:conclusions}
\vspace{-5pt}
We introduced \ourtaskshort, a task for structured video similarity estimation, and \ourdataset, a novel human-annotated dataset of video pairs designed to evaluate video similarity along core semantic concepts, covering the widest range of domains to date.
We extensively benchmark models, including ten state-of-the-art LMMs, on \ourdataset, revealing substantial variability in how they align with human judgments,  for the first time, exposing their biases toward particular concepts.
By enabling concept-conditioned evaluation and retrieval, \ourtaskshort offers a path towards more interpretable, controllable, and user-aligned video understanding systems.

\textbf{Limitations and Future Work.}
In this work, we focused on a small set of general-purpose concepts to enable consistent and interpretable comparisons across diverse video domains.
While this choice, grounded in cognitive science, supports broad applicability, it also introduces a limitation: the current concept set may not capture all domain-specific or fine-grained aspects of video similarity.
We believe the same framework can be extended to incorporate additional concepts, enabling richer analyses in specialized settings.
Moreover, \ourdataset currently includes a relatively modest number of video pairs. Although limited in scale, this is a consequence of prioritizing annotation quality over quantity.
Expanding the dataset, provided quality standards are maintained, would further increase its value and support broader evaluations.

\clearpage
\bibliographystyle{plainnat}  
\bibliography{main} 
\clearpage
\newpage
\appendix
\section*{\Large Appendix}

In this Appendix, we first provide further details about the interface used for the annotation of \ourdataset in Sec.~\ref{sec:annotation_interface}. In Sec.~\ref{sec:dataset_analysis}, we provide additional statistics and insights about our proposed dataset. In Sec.~\ref{sec:llm_prompts} we show the prompts used for the evaluation of the LMMs and in Sec.~\ref{sec:implementation_details} we provide additional implementation details. In Sec.~\ref{sec:additional_experiments} we extend the experiments provided in the main paper, and in Sec.~\ref{sec:qualitative_examples} we show additional qualitative examples from \ourdataset. Finally, in Sec.~\ref{sec:broader_impacts} we discuss the broader impacts of our work, in Sec.~\ref{sec:review_board} we declare that our institutional review board approved the present study, and list the licenses and URLs for all external assets used in Section~\ref{sec:assets}.

\section{Annotation interface}\label{sec:annotation_interface}

In this section, we provide all the details about the human annotation process. 
We show screenshots of our interface, a matrix outlining the scoring guidelines, and information about annotator screening procedures and compensation.

\begin{figure}[ht]
    \centering
    \includegraphics[width=1\linewidth]{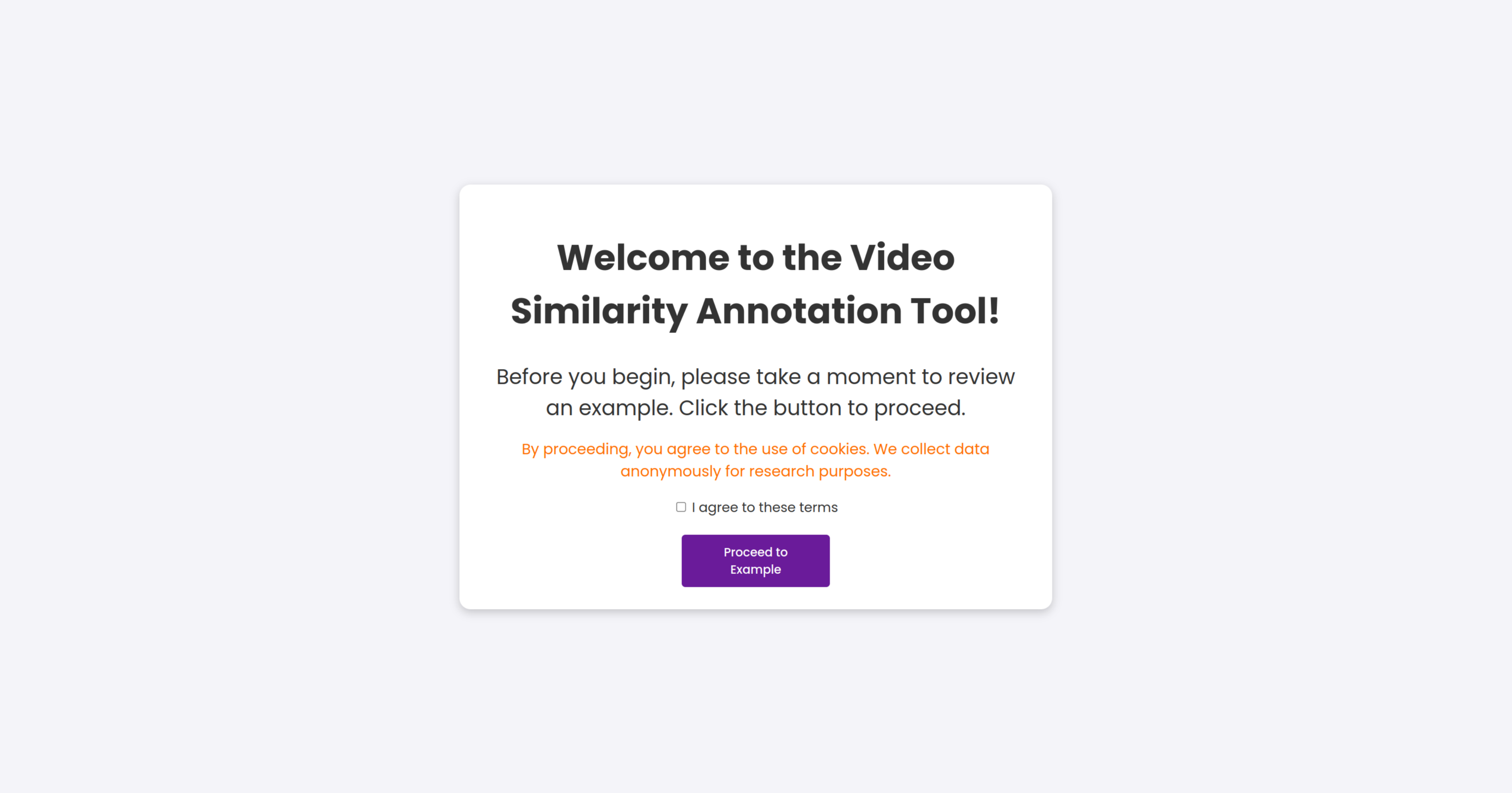}
    \caption{\textbf{Welcome page of the annotation interface.}}
    \label{fig:annotation_welcome}
\end{figure}

\begin{figure}[ht]
    \centering
    \includegraphics[width=1\linewidth]{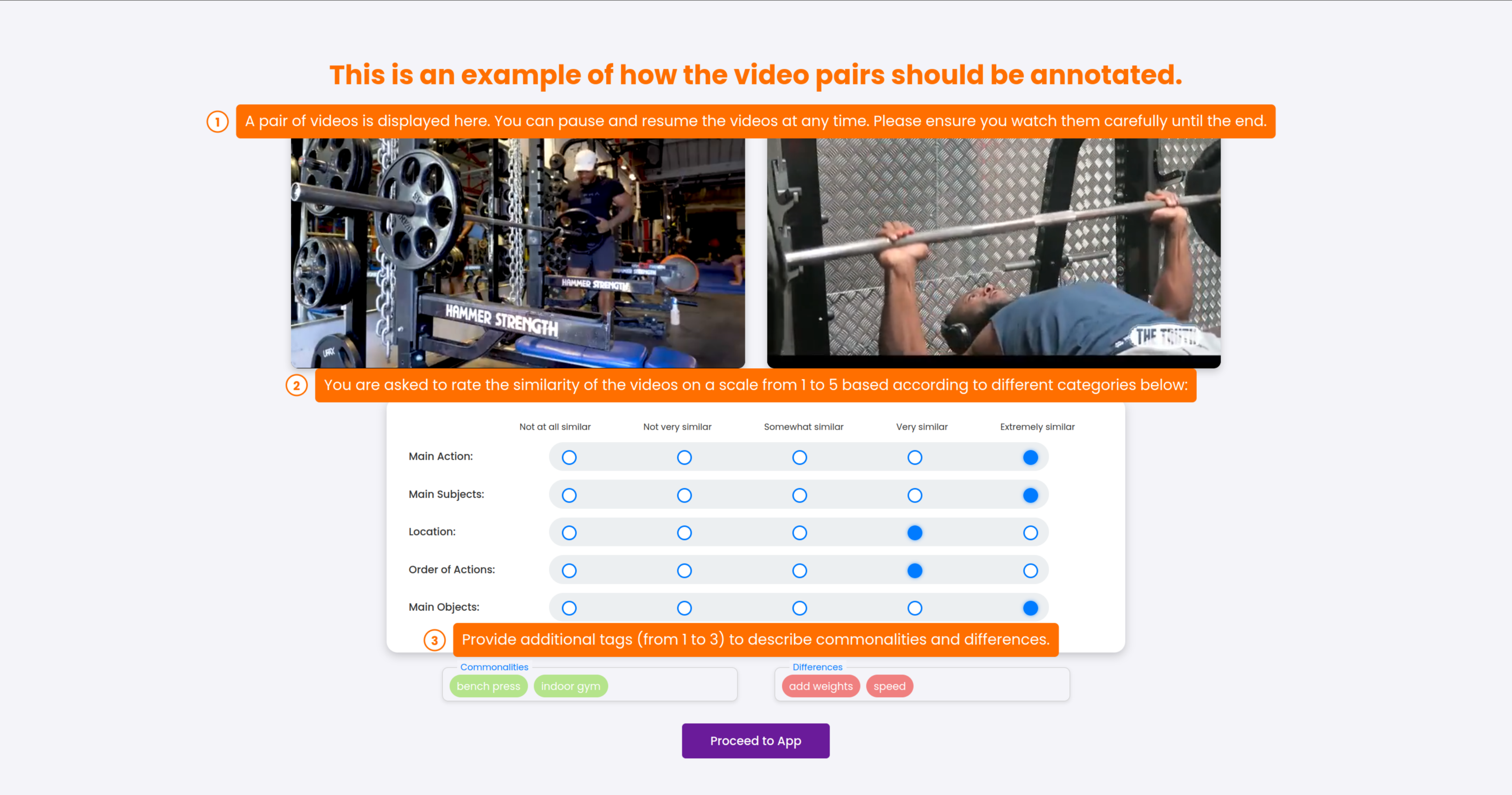}
    \caption{\textbf{Provided example in the annotation interface.}}
    \label{fig:annotation_example}
\end{figure}

\begin{figure}[ht]
    \centering
    \includegraphics[width=1\linewidth]{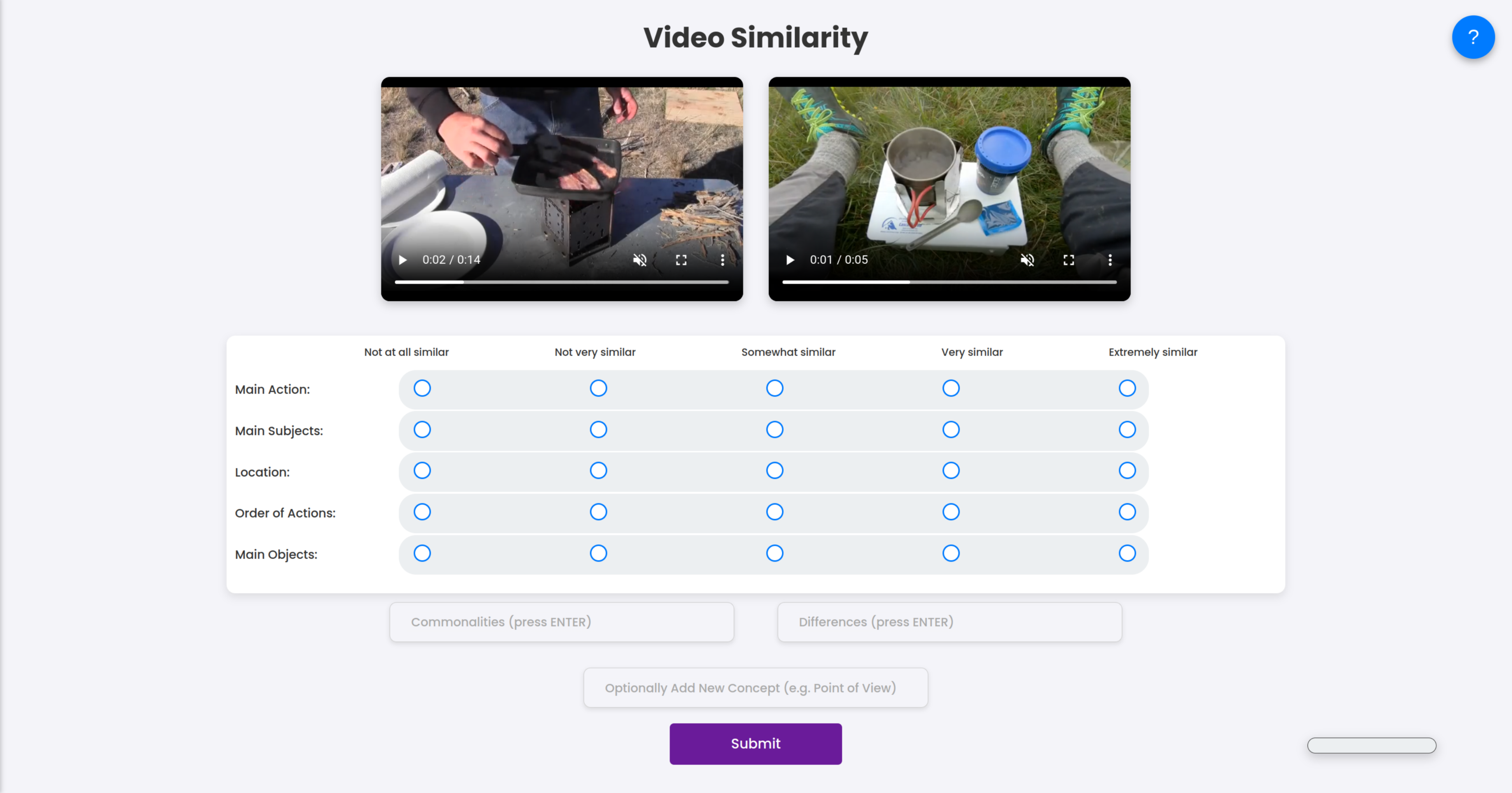}
    \caption{\textbf{Annotation interface.}}
    \label{fig:annotation_task}
\end{figure}

As a first step, annotators are asked to accept the use of cookies (see Fig.~\ref{fig:annotation_welcome}). We do not collect any personally identifiable information, such as email addresses, names, or nicknames. However, we keep track of the reviewed samples to ensure that the same annotation pairs are not shown to the same annotator multiple times.
Next, we present an example to illustrate how to complete the task correctly (see Fig.~\ref{fig:annotation_example}).
Finally, Fig.~\ref{fig:annotation_task} shows the main annotation interface used to carry out the task.
The annotation interface includes task instructions, which can be viewed at any time, as well as embedded guidelines that appear when hovering over relevant buttons. 
This appendix also contains a video demonstrating the interface in use to provide a clearer understanding of the annotation process.
A scoring matrix summarizing these guidelines is also displayed for reference in  Tab.~\ref{tab:scoring_guideline}.

Annotators are screened to ensure they are college-educated and reside in English-speaking countries. 
They are compensated at a rate of \pounds8~per hour.

\setlength{\arrayrulewidth}{0.3mm}
\renewcommand{\arraystretch}{1.8}
\newcolumntype{Y}{>{\raggedright\arraybackslash}X}
\begin{table}[ht]
\centering
\scriptsize
\caption{\textbf{Scoring guidelines provided in the annotation interface.}}
\label{tab:scoring_guideline}
\rowcolors{2}{blue!10}{white}
\begin{tabularx}{\textwidth}{|>{\columncolor{blue!10}}l|Y|Y|Y|Y|Y|}
\hline
\rowcolor{blue!30}
\textbf{Concepts} & \textbf{1} & \textbf{2} & \textbf{3} & \textbf{4} & \textbf{5} \\
\hline
\texttt{Main Action} & Completely different types of action (\eg, cooking and sports domains) & Actions are different but related (\eg, both sports) & Actions are similar with significant differences in how they are carried out & Actions are similar with slight differences in how they are carried out & No difference in the main action performed \\
\hline
\texttt{Main Subjects} & Totally different cardinality, gender and age (\eg, a man and a group of women) & Different cardinality but some similar in gender or age (\eg, a man and a group of men) & Same cardinality but significant difference in age or gender (\eg, a man and a boy) & Same cardinality but minor difference in age (\eg, a young male and a boy) & Extremely similar (\eg, two young women in both) \\
\hline
\texttt{Location} & Completely different places & Different but similar type of place (\eg, indoor and outdoor soccer field) & Similar locations still with significant differences & Similar locations with slight differences & Extremely similar (\eg, rooms with a similar size and layout) \\
\hline
\texttt{Order of Actions} & Completely different, or no sub-actions are in common & The order of actions has significant changes, and some missing steps & Some actions are reordered, with a few steps missing, but key steps remain intact & Minor changes in the order of actions & Actions occur in the same order, without missing steps \\
\hline
\texttt{Main Objects} & Main objects are entirely different & A few main objects are shared, but most are different & Most main objects are similar, with some not in common & Minor differences in main objects (\eg, color or shape) & Main objects are the same or extremely similar \\
\hline
\end{tabularx}
\end{table}

\section{Additional dataset analysis}\label{sec:dataset_analysis}

\textbf{Annotation quality assessment.}
We assess annotation quality by measuring inter-annotator agreement (IAA) using Krippendorff’s $\alpha$~\cite{Krippendorff1980}. 
After removing low-quality samples through our end-stage curation process, the resulting $\alpha$ scores for each concept are: \textit{main subjects} (0.244), \textit{main action} (0.322), \textit{location} (0.361), \textit{main objects} (0.288), and \textit{order of actions} (0.319).
On average, this curation step improved inter-annotator agreement by 17.5\%, highlighting the effectiveness of our quality control process.

\begin{figure}[!ht]
    \centering
    \includegraphics[width=0.9\linewidth]{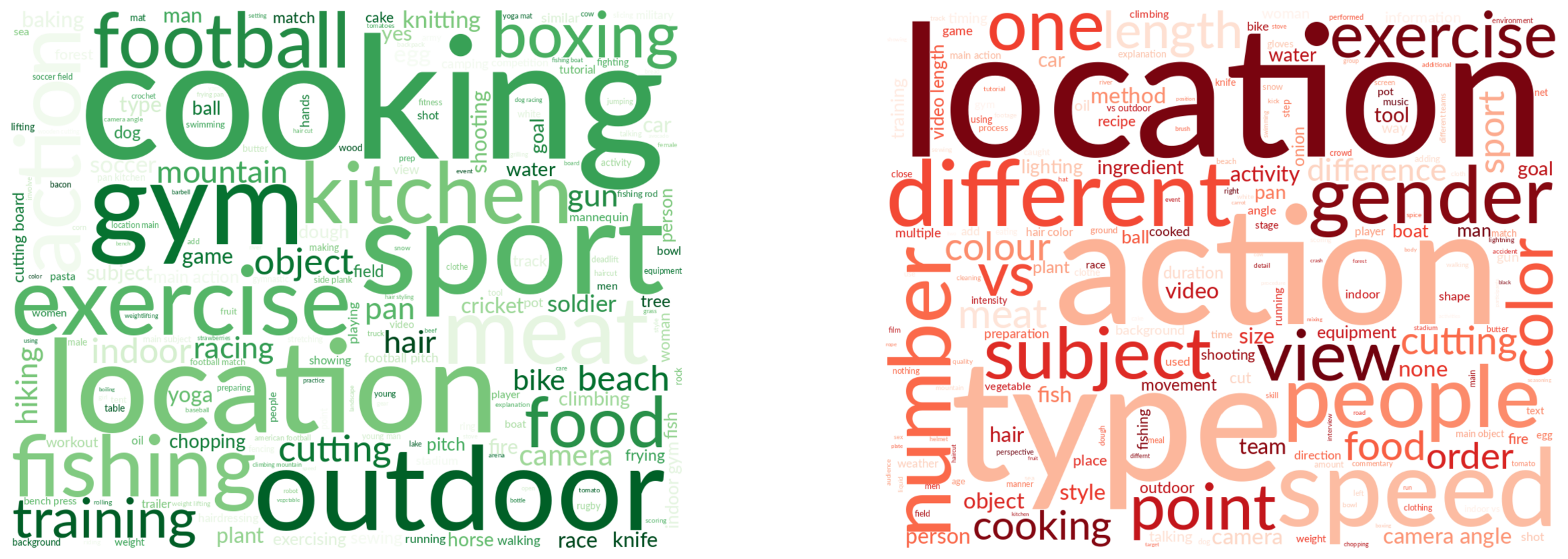}
    \caption{\textbf{Wordclouds of collected tags.} Visualization of tags for \inlineColorbox{CustomGreen2!50}{similarities} and \inlineColorbox{CustomRed}{differences}, displayed as wordclouds.}
    \label{fig:concept-wordcloud}
\end{figure}

\textbf{Statistics on tags.}
We obtain, on average, 15 textual tags for each video pair. Figure~\ref{fig:concept-wordcloud} shows wordclouds of prominent tags for similarities and differences. From these wordclouds, we observe that most tags relate to actions (\eg, \textit{gym}, \textit{cooking}, \textit{boxing}, \textit{fishing}), objects (\eg, \textit{food}, \textit{bike}, \textit{gun}), subjects (\eg, \textit{people}, \textit{gender}), or locations (\eg, \textit{location}, \textit{beach}, \textit{mountain}). 
A smaller portion of tags describe aspects of time, such as \textit{order} and \textit{speed}.

\begin{figure}[h!]
    \centering
    \includegraphics[width=\linewidth]{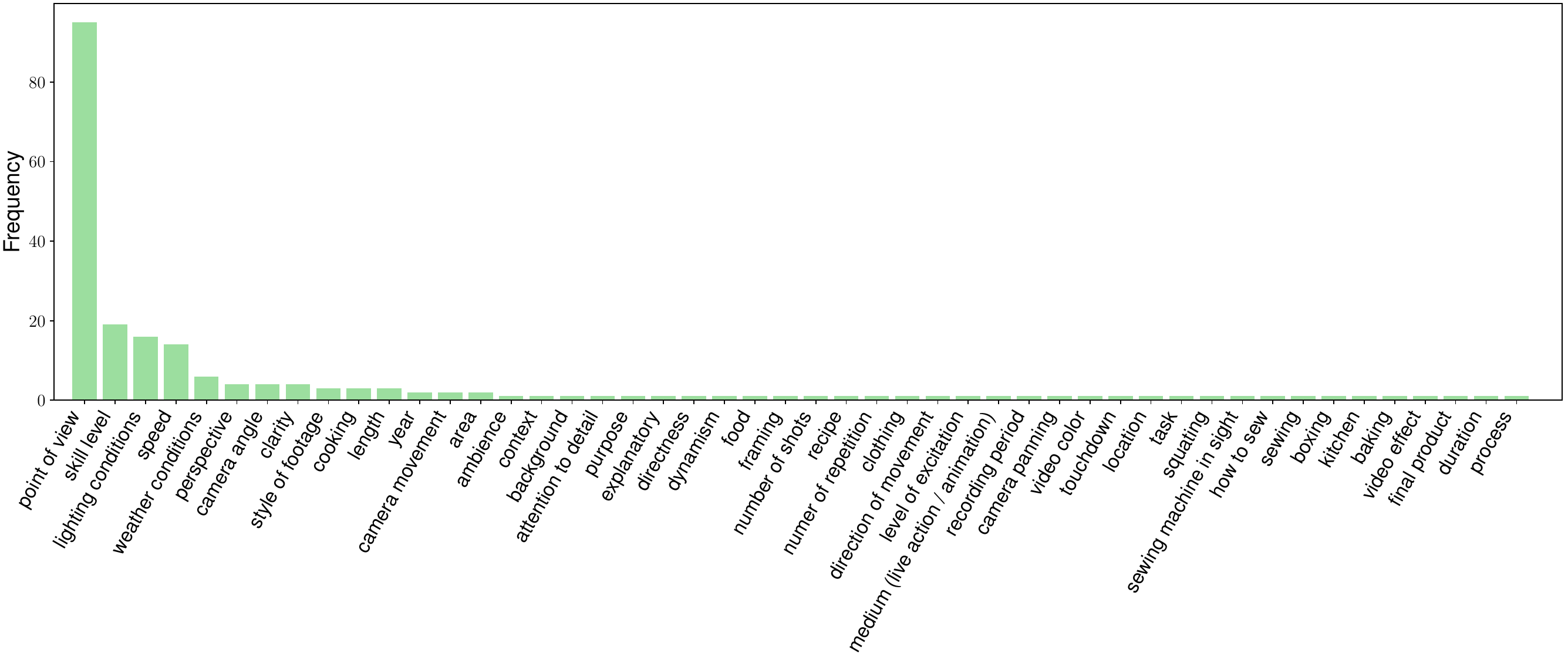}
    \caption{\textbf{User-defined concepts.}}
    \label{fig:user_defined_concepts}
\end{figure}

\textbf{User-defined concepts.}
Figure~\ref{fig:user_defined_concepts} reports the frequencies of concepts proposed by annotators beyond the predefined set. 
The most commonly added concepts include point of view, skill level, and lighting condition. 
However, even the most frequently mentioned concept appears in fewer than $16\%$ of video pairs, indicating that such additions are relatively sparse.
These user-defined concepts highlight directions for further development of the concept taxonomy.

\begin{figure}[ht]
    \begin{minipage}[t]{0.45\linewidth}
        \centering
        \includegraphics[width=\linewidth]{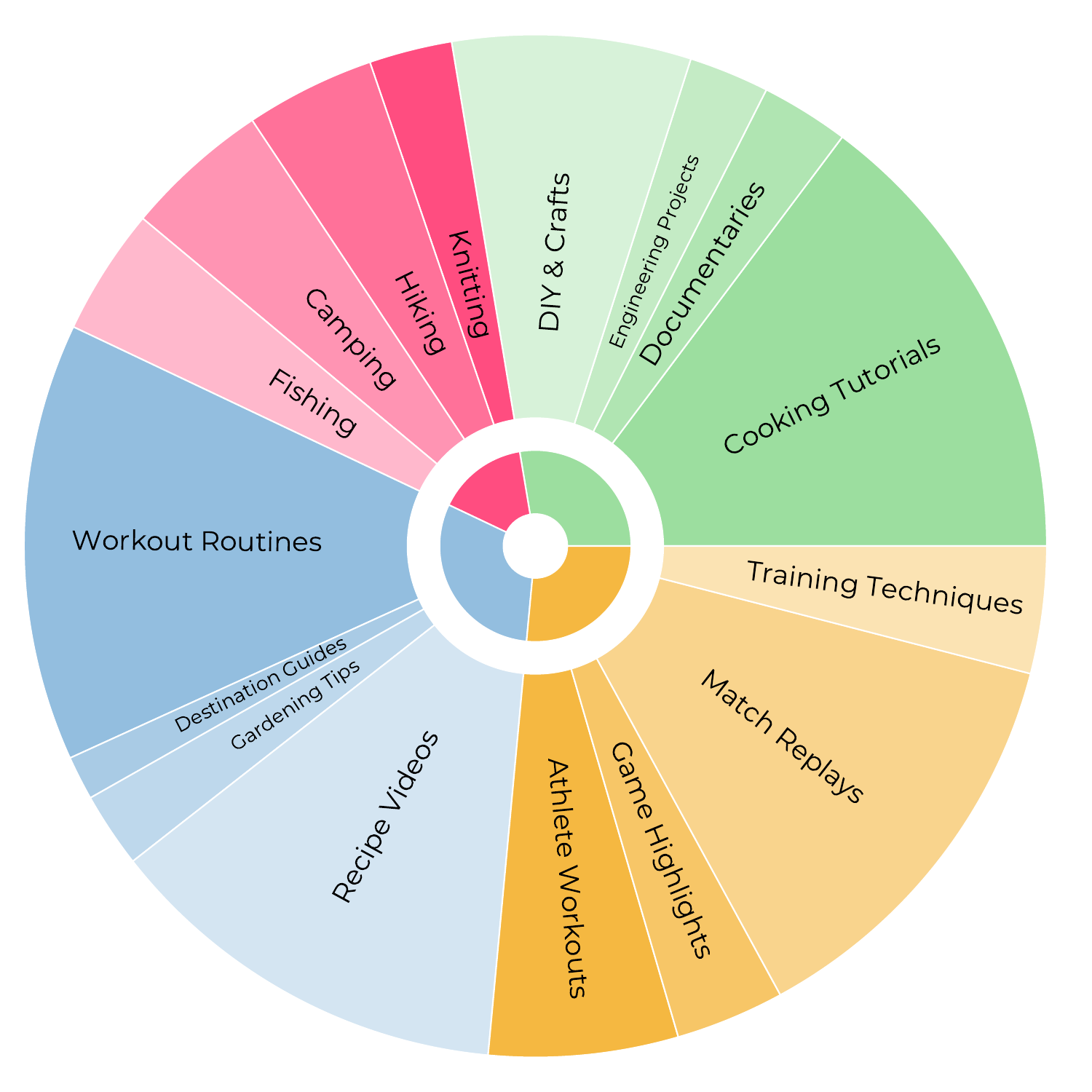}
        \caption{\textbf{Video distribution per domain in \ourdataset.}}
        \label{fig:video_distribution}
    \end{minipage}
    \hfill
    \begin{minipage}[t]{0.45\linewidth}
        \centering
        \includegraphics[width=\linewidth]{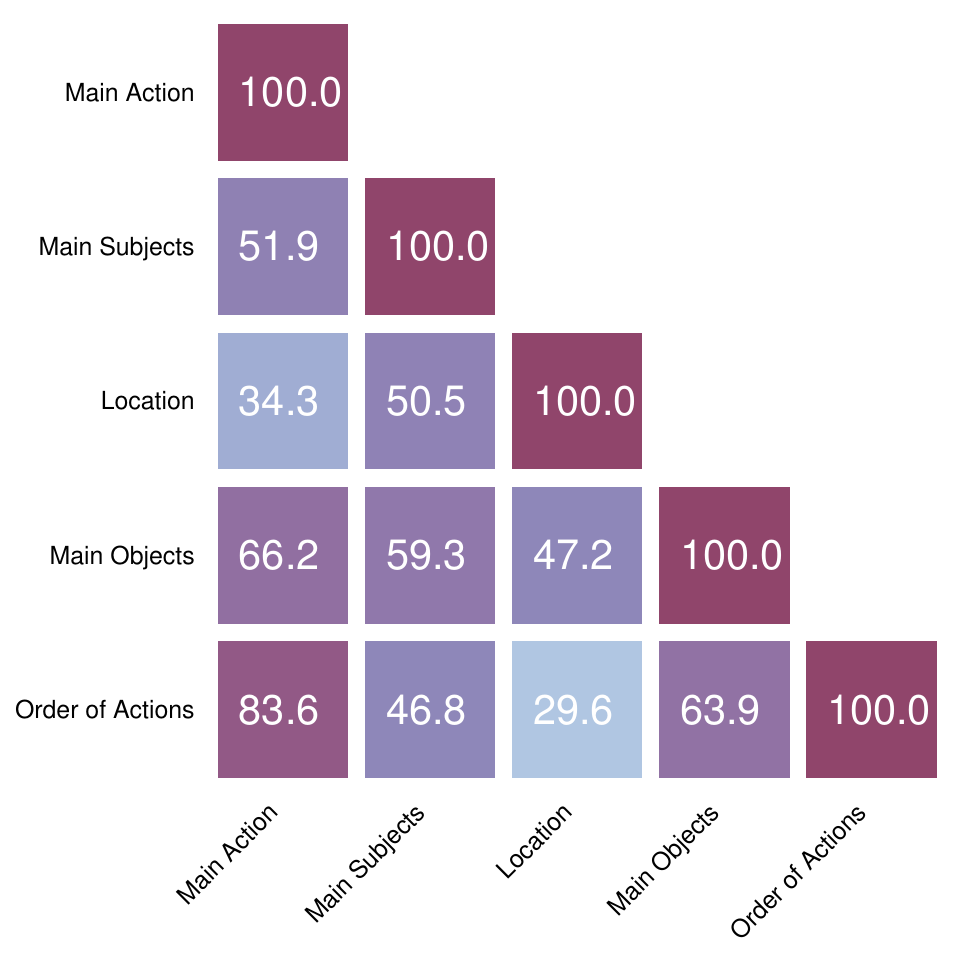}
        \caption{\textbf{Concept inter-correlation.} We report Spearman's $\rho$ correlation ($\times100$) between different concepts across the dataset.}
        \label{fig:gt_inter_correlation}
    \end{minipage}
\end{figure}

\textbf{Domains}
The videos in \ourdataset span 16 distinct categories, as shown in Fig.~\ref{fig:video_distribution}. 
The category names are retained from the source dataset~\cite{FineVideo}, where they are grouped into top-level categories. These include \textit{Sports}, which encompasses \textit{Athlete Workouts, Training Techniques, Game Highlights, and Match Replays}; \textit{Lifestyle}, which includes \textit{Recipe Videos, Gardening Tips, Workout Routines, and Destination Guides}; \textit{Education}, consisting of \textit{Engineering Projects, Cooking Tutorials, Documentaries, and DIY \& Crafts}; and finally, \textit{Hobbies \& Interests}, which covers \textit{Knitting, Hiking, Camping, and Fishing}.

\begin{figure}[h!]
    \centering
    \includegraphics[width=1\linewidth]{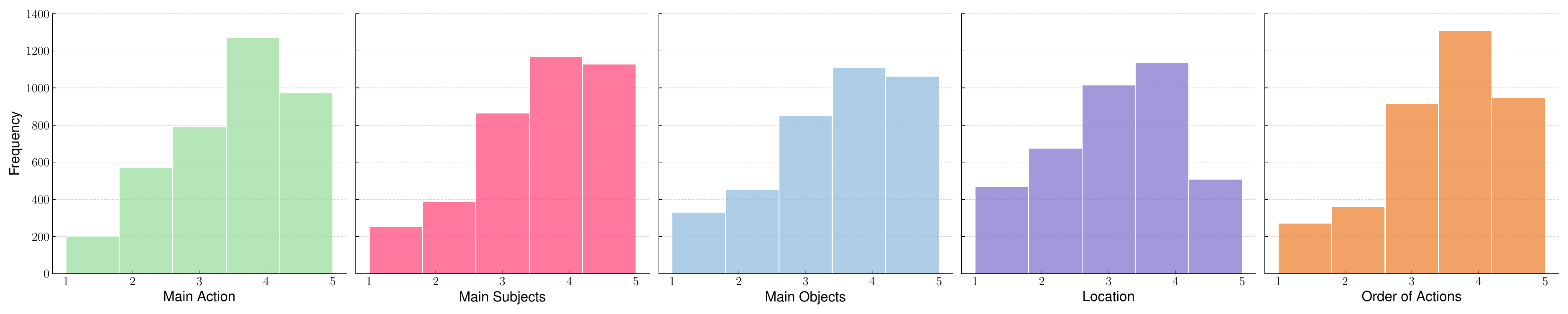}
    \caption{\textbf{Frequency of scores assigned by human annotators, divided by concepts.}}
    \label{fig:score_frequency}
\end{figure}

\textbf{Statistics on scores.} 
Figure~\ref{fig:score_frequency} reports the frequency distribution of the collected similarity scores for each concept. The distributions vary across concepts, with \textit{main objects} exhibiting the highest number of scores equal to 1. 
Furthermore, Figure~\ref{fig:gt_inter_correlation} shows the inter-correlation matrix between the scores assigned to different concepts. These results highlight that, according to human annotators, most visual concepts are not strongly correlated. The only notable exception is the pair \textit{main action} and \textit{order of action}, which exhibit a correlation higher than~$80$.

\section{Prompts for LMMs evaluation}\label{sec:llm_prompts}

Figure~\ref{fig:prompt} illustrates the prompts used to evaluate LMMs on \ourtaskshort. 
These prompts, for all the concepts, are used in the evaluation setup described in Sec.~\ref{sec:consim}.

\begin{figure}[h!]
\promptbox{System Prompt}{
\texttt{You are an AI designed to compare two videos based on the visual concept of \textcolor{purple}{<concept>}}.\newline

\texttt{The input consists of a sequence of concatenated frames: the first half represents Video 1, and the second half represents Video 2.}

\texttt{Your task is to evaluate how similar these two videos are with respect to the concept \textcolor{purple}{<concept>}.}

\texttt{Output a single similarity score between 1 and 5, where 1 means completely different and 5 means perfectly similar in terms of \textcolor{purple}{<concept>}.}

\texttt{Do not explain your reasoning. Only output the numerical score.}}
\caption{}
\label{fig:prompt}
\end{figure}

\section{Implementation details}\label{sec:implementation_details}

Regarding the specific models used in Sec~\ref{sec:globalsim}, we use DINOv2~\cite{oquab2024dinov}, CLIP~\cite{radford2021learning}, and InternVideo~\cite{wang2022internvideo} with a ViT-L/14 backbone, VideoMAE~\cite{NEURIPS2022_416f9cb3} with ViT-L, and Sentence-BERT~\cite{reimers2019sentencebert} with the MiniLM-L6 (\ie, \texttt{all-MiniLM-L6-v2} variant), which is a widely adopted model for text-to-text semantic similarity tasks.

\section{Additional experiments}\label{sec:additional_experiments}

Table~\ref{tab:conditioned_single_frame_delta} reports the performance differences when using a single frame per video instead of the full sequence of frames. 
It shows detailed results for all the concepts, with negative values indicating degraded alignment with human judgments. 
Tab.~\ref{tab:concept_based_corr_1frame} reports the extended results across varying numbers of frames per video, illustrating how human-alignment scales with temporal input.

\begin{table}[ht!]
\def\arraystretch{1.15}
\centering
\caption{\textbf{Alignment performance delta when using a single frame per video.} We report the difference ($\Delta$) in Spearman's $\rho$ and Kendall's $\tau$ correlations with human judgement ($\times100$), computed as the change in performance when models receive a single frame per video instead of $N$ frames. A negative value indicates performance degradation with fewer frames. A dash (–) indicates that the correlation could not be computed because the model outputs constant predictions.}
\scriptsize
\begin{tabularx}{\textwidth}{
>{\raggedright\arraybackslash}p{2.5cm}>{\centering\arraybackslash}X>{\centering\arraybackslash}X>{\centering\arraybackslash}X>{\centering\arraybackslash}X>{\centering\arraybackslash}X>{\centering\arraybackslash}X>{\centering\arraybackslash}X>{\centering\arraybackslash}X>{\centering\arraybackslash}X>{\centering\arraybackslash}X}
\toprule
 & \multicolumn{2}{c}{\texttt{Main Action}} & \multicolumn{2}{c}{\texttt{Main Subjects}} & \multicolumn{2}{c}{\texttt{Main Objects}} & \multicolumn{2}{c}{\texttt{Location}} & \multicolumn{2}{c}{\texttt{Order of Actions}} \\
Model & $\Delta\rho$ & $\Delta\tau$ & $\Delta\rho$ & $\Delta\tau$ & $\Delta\rho$ & $\Delta\tau$ & $\Delta\rho$ & $\Delta\tau$ & $\Delta\rho$ & $\Delta\tau$ \\
\midrule
mPLUG-Owl3-7B~\cite{ye2025mplugowl} & \ccc{-6.45} & \ccc{-5.13} & \ccc{-1.86} & \ccc{-1.35} & \ccc{-10.89} & \ccc{-8.85} & \ccc{-11.14} & \ccc{-8.89} & \ccc{+0.57} & \ccc{+0.78} \\
LLaVA-OV-0.5B~\cite{li2025llavaonevision} & \ccc{-3.84} & \ccc{-3.11} & \ccc{+0.62} & \ccc{+0.24} & \ccc{+3.50} & \ccc{+2.79} & \ccc{-6.11} & \ccc{-4.96} & \ccc{+4.39} & \ccc{+3.81} \\
LLaVA-OV-7B~\cite{li2025llavaonevision} & \ccc{-19.40} & \ccc{-16.24} & \ccc{-11.37} & \ccc{-9.05} & \ccc{-20.93} & \ccc{-17.69} & \ccc{-14.88} & \ccc{-11.86} & \ccc{-15.22} & \ccc{-12.36} \\
LLaVA-Video-7B~\cite{zhang2024video} & \ccc{-13.56} & \ccc{-11.96} & \ccc{-11.09} & \ccc{-9.65} & \ccc{-1.96} & \ccc{-2.26} & \ccc{-10.65} & \ccc{-9.45} & \ccc{-15.81} & \ccc{-13.23} \\
Qwen2.5-VL-3B~\cite{bai2025qwen2} & \ccc{-26.80} & \ccc{-22.22} & \ccc{-10.96} & \ccc{-8.64} & \ccc{-17.98} & \ccc{-14.92} & \ccc{-13.03} & \ccc{-10.80} & \ccc{-} & \ccc{-} \\
Qwen2.5-VL-7B~\cite{bai2025qwen2} & \ccc{-29.57} & \ccc{-24.39} & \ccc{-16.45} & \ccc{-13.53} & \ccc{-17.92} & \ccc{-14.75} & \ccc{-21.96} & \ccc{-17.80} & \ccc{-22.81} & \ccc{-18.75} \\
InternVL2.5-4B~\cite{chen2024expanding} & \ccc{+14.88} & \ccc{+11.55} & \ccc{+5.03} & \ccc{+3.87} & \ccc{+17.84} & \ccc{+13.88} & \ccc{+5.82} & \ccc{+4.54} & \ccc{+15.65} & \ccc{+11.93} \\
InternVL2.5-8B~\cite{chen2024expanding} & \ccc{-8.91} & \ccc{-7.38} & \ccc{-1.59} & \ccc{-1.71} & \ccc{-1.73} & \ccc{-1.43} & \ccc{+8.70} & \ccc{+6.65} & \ccc{-2.33} & \ccc{-2.06} \\
InternVL3-8B~\cite{zhu2025internvl3} & \ccc{-0.59} & \ccc{-0.09} & \ccc{-3.56} & \ccc{-2.26} & \ccc{+2.93} & \ccc{+3.01} & \ccc{+2.12} & \ccc{+2.10} & \ccc{+1.67} & \ccc{+1.56} \\
Gemini-2.0-Flash~\cite{team2023gemini} & \ccc{-3.57} & \ccc{-2.06} & \ccc{-0.93} & \ccc{-0.62} & \ccc{+9.85} & \ccc{+8.62} & \ccc{+7.01} & \ccc{+5.10} & \ccc{-6.86} & \ccc{-4.59} \\
\bottomrule
\end{tabularx}
\label{tab:conditioned_single_frame_delta}
\end{table}

\begin{table}[ht!]
\def\arraystretch{1.15}
\centering
\caption{\textbf{Alignment performance of LMMs on \ourdataset at varying number of frames.} We report Spearman's $\rho$ and Kendall's $\tau$ correlations with human judgment ($\times100$). \textit{Frames} indicates the number of frames per every video in the pair.  A dash (–) indicates that the correlation could not be computed because the model outputs constant predictions.}
\scriptsize
\begin{tabularx}{\textwidth}{
>{\raggedright\arraybackslash}p{1.8cm}>{\centering\arraybackslash}X>{\centering\arraybackslash}X>{\centering\arraybackslash}X>{\centering\arraybackslash}X>{\centering\arraybackslash}X>{\centering\arraybackslash}X>{\centering\arraybackslash}X>{\centering\arraybackslash}X>{\centering\arraybackslash}X>{\centering\arraybackslash}X>{\centering\arraybackslash}X}
\toprule
Model & Frames & \multicolumn{2}{c}{\texttt{Main Action}} & \multicolumn{2}{c}{\texttt{Main Subjects}} & \multicolumn{2}{c}{\texttt{Main Objects}} & \multicolumn{2}{c}{\texttt{Location}} & \multicolumn{2}{c}{\texttt{Order of Actions}} \\
 & & $\rho$ & $\tau$ & $\rho$ & $\tau$ & $\rho$ & $\tau$ & $\rho$ & $\tau$ & $\rho$ & $\tau$ \\
\midrule
mPLUG-Owl37B & 16 & 30.64 & 25.12 & 20.59 & 16.84 & 28.53 & 23.47 & 21.00 & 17.00 & 23.11 & 18.74 \\
mPLUG-Owl37B & 8 & 31.18 & 25.62 & 23.15 & 18.98 & 29.05 & 23.95 & 22.02 & 17.98 & 21.43 & 17.51 \\
mPLUG-Owl37B & 4 & 32.82 & 27.09 & 23.19 & 19.23 & 33.67 & 27.92 & 14.14 & 11.64 & 23.73 & 19.56 \\
mPLUG-Owl37B & 2 & 29.32 & 24.20 & 18.69 & 15.49 & 25.04 & 20.77 & 8.12 & 6.67 & 23.81 & 19.62 \\
mPLUG-Owl37B & 1 & 24.19 & 19.99 & 18.73 & 15.49 & 17.64 & 14.62 & 9.86 & 8.11 & 23.68 & 19.52 \\
\midrule
LLaVA-OV-0.5B & 16 & 1.95 & 1.54 & -5.05 & -3.91 & -4.00 & -3.20 & 5.66 & 4.59 & 1.30 & 0.90 \\
LLaVA-OV-0.5B & 8 & -4.04 & -3.32 & -8.69 & -6.87 & 0.09 & 0.06 & -1.06 & -0.81 & -0.32 & -0.30 \\
LLaVA-OV-0.5B & 4 & 3.75 & 3.09 & 6.52 & 5.36 & 10.60 & 8.74 & 4.73 & 3.83 & 7.02 & 5.78 \\
LLaVA-OV-0.5B & 2 & 6.12 & 5.01 & -2.59 & -2.14 & 0.35 & 0.29 & -1.58 & -1.33 & 5.66 & 4.66 \\
LLaVA-OV-0.5B & 1 & -1.89 & -1.57 & -4.43 & -3.67 & -0.50 & -0.41 & -0.45 & -0.37 & 5.69 & 4.71 \\
\midrule
LLaVA-OV-7B & 16 & 51.76 & 42.48 & 48.43 & 39.14 & 58.64 & 48.17 & 58.94 & 47.72 & 41.02 & 33.31 \\
LLaVA-OV-7B & 8 & 47.61 & 38.94 & 47.30 & 38.64 & 50.23 & 41.36 & 57.00 & 46.10 & 43.46 & 35.49 \\
LLaVA-OV-7B & 4 & 44.39 & 35.87 & 39.84 & 32.28 & 48.68 & 40.02 & 57.39 & 46.55 & 37.51 & 30.54 \\
LLaVA-OV-7B & 2 & 35.81 & 28.96 & 40.51 & 33.04 & 45.06 & 36.93 & 50.33 & 40.90 & 32.60 & 26.72 \\
LLaVA-OV-7B & 1 & 32.36 & 26.24 & 37.06 & 30.09 & 37.71 & 30.48 & 44.06 & 35.86 & 25.80 & 20.95 \\
\midrule
LLaVA-Video-7B & 16 & 44.17 & 36.53 & 39.81 & 32.81 & 45.85 & 37.96 & 55.96 & 46.33 & 41.25 & 34.05 \\
LLaVA-Video-7B & 8 & 44.86 & 36.97 & 35.08 & 28.85 & 43.71 & 36.12 & 54.84 & 45.16 & 39.43 & 32.59 \\
LLaVA-Video-7B & 4 & 46.50 & 38.42 & 35.88 & 29.57 & 43.76 & 36.12 & 53.98 & 44.54 & 39.86 & 32.95 \\
LLaVA-Video-7B & 2 & 37.79 & 30.95 & 32.00 & 26.33 & 40.88 & 33.55 & 48.22 & 39.33 & 33.68 & 27.81 \\
LLaVA-Video-7B & 1 & 30.61 & 24.57 & 28.72 & 23.16 & 43.89 & 35.70 & 45.31 & 36.88 & 25.44 & 20.82 \\
\midrule
Qwen2.5-VL-3B & 16 & 21.84 & 18.10 & 8.62 & 6.70 & 15.20 & 12.61 & 13.44 & 11.14 & 12.79 & 10.60 \\
Qwen2.5-VL-3B & 8 & 23.34 & 19.35 & 16.62 & 13.75 & 20.20 & 16.75 & 14.27 & 11.76 & 8.37 & 6.91 \\
Qwen2.5-VL-3B & 4 & 20.56 & 17.06 & 12.65 & 10.48 & 7.09 & 5.88 & 13.28 & 11.00 & 5.86 & 4.85 \\
Qwen2.5-VL-3B & 2 & 9.76 & 8.09 & 7.51 & 6.22 & 8.11 & 6.73 & 4.32 & 3.57 & - & - \\
Qwen2.5-VL-3B & 1 & -4.96 & -4.12 & -2.34 & -1.94 & -2.78 & -2.31 & 0.41 & 0.34 & - & - \\
\midrule
Qwen2.5-VL-7B & 16 & 37.88 & 31.28 & 17.53 & 14.43 & 26.97 & 22.26 & 23.63 & 19.17 & 23.85 & 19.61 \\
Qwen2.5-VL-7B & 8 & 29.13 & 23.92 & 16.02 & 13.21 & 24.23 & 20.01 & 19.59 & 16.02 & 20.24 & 16.64 \\
Qwen2.5-VL-7B & 4 & 26.47 & 21.83 & 19.53 & 16.08 & 19.75 & 16.35 & 18.29 & 15.03 & 22.88 & 18.89 \\
Qwen2.5-VL-7B & 2 & 17.84 & 14.76 & 13.99 & 11.58 & 10.85 & 9.00 & 7.34 & 6.08 & 16.01 & 13.23 \\
Qwen2.5-VL-7B & 1 & 8.31 & 6.89 & 1.08 & 0.90 & 9.05 & 7.51 & 1.67 & 1.37 & 1.04 & 0.86 \\
\midrule
InternVL2\_5-4B & 16 & 13.23 & 10.13 & 16.71 & 12.89 & 14.99 & 11.71 & 13.93 & 10.71 & 9.88 & 7.54 \\
InternVL2\_5-4B & 8 & 17.31 & 13.30 & 19.95 & 15.51 & 21.42 & 17.05 & 22.18 & 17.13 & 9.98 & 7.68 \\
InternVL2\_5-4B & 4 & 27.58 & 21.07 & 28.47 & 21.70 & 31.76 & 24.63 & 21.47 & 16.12 & 19.77 & 14.99 \\
InternVL2\_5-4B & 2 & 27.98 & 21.34 & 22.97 & 17.44 & 29.00 & 22.22 & 22.37 & 17.10 & 22.63 & 17.16 \\
InternVL2\_5-4B & 1 & 28.11 & 21.68 & 21.74 & 16.76 & 32.83 & 25.59 & 19.75 & 15.25 & 25.53 & 19.47 \\
\midrule
InternVL2\_5-8B & 16 & 28.70 & 22.42 & 28.60 & 22.08 & 25.06 & 19.40 & 19.64 & 15.18 & 18.15 & 14.07 \\
InternVL2\_5-8B & 8 & 22.93 & 17.68 & 27.35 & 20.99 & 26.67 & 20.65 & 32.21 & 24.96 & 25.20 & 19.31 \\
InternVL2\_5-8B & 4 & 28.53 & 21.95 & 30.41 & 23.20 & 32.27 & 24.89 & 28.17 & 21.80 & 22.98 & 17.46 \\
InternVL2\_5-8B & 2 & 25.30 & 19.43 & 27.19 & 20.53 & 27.87 & 21.46 & 31.65 & 24.36 & 21.87 & 16.46 \\
InternVL2\_5-8B & 1 & 19.79 & 15.04 & 27.01 & 20.37 & 23.33 & 17.97 & 28.34 & 21.83 & 15.82 & 12.01 \\
\midrule
InternVL3-8B & 16 & 40.69 & 31.31 & 36.54 & 27.80 & 42.50 & 32.88 & 45.47 & 34.98 & 32.74 & 24.88 \\
InternVL3-8B & 8 & 45.68 & 35.26 & 41.74 & 31.86 & 47.25 & 36.79 & 49.04 & 38.08 & 31.98 & 24.25 \\
InternVL3-8B & 4 & 48.24 & 37.23 & 42.11 & 31.91 & 49.65 & 38.65 & 53.78 & 41.75 & 32.51 & 24.69 \\
InternVL3-8B & 2 & 41.84 & 32.43 & 34.02 & 26.07 & 41.49 & 32.33 & 52.05 & 40.44 & 30.64 & 23.52 \\
InternVL3-8B & 1 & 40.10 & 31.22 & 32.98 & 25.54 & 45.43 & 35.89 & 47.59 & 37.08 & 34.41 & 26.44 \\
\bottomrule
\end{tabularx}
\label{tab:concept_based_corr_1frame}
\end{table}

\section{Qualitative examples}\label{sec:qualitative_examples}

\begin{figure}
    \centering

    \begin{subfigure}{\linewidth}
        \includegraphics[width=\linewidth]{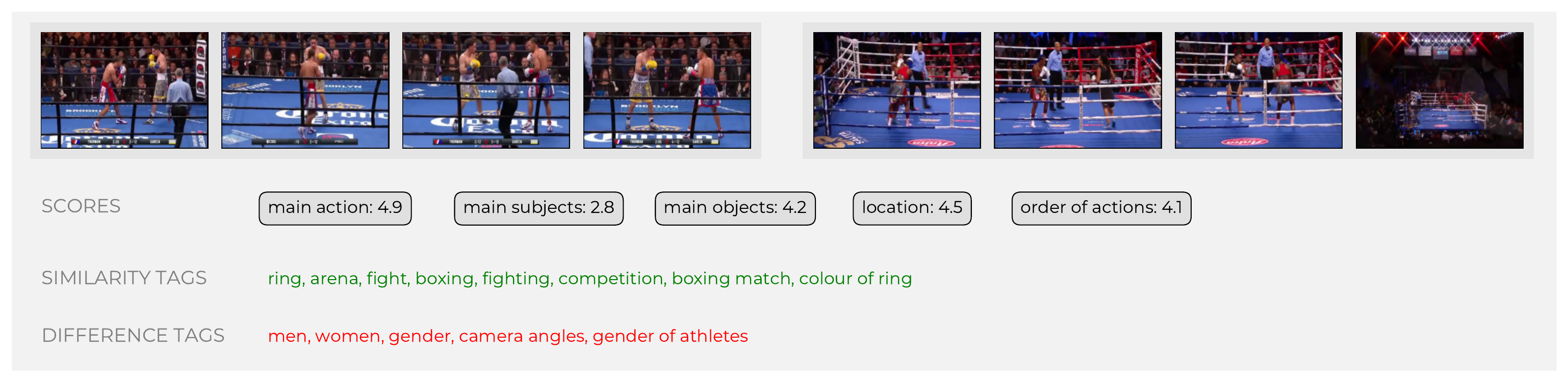}
        \caption{}
    \end{subfigure}

    \begin{subfigure}{\linewidth}
        \includegraphics[width=\linewidth]{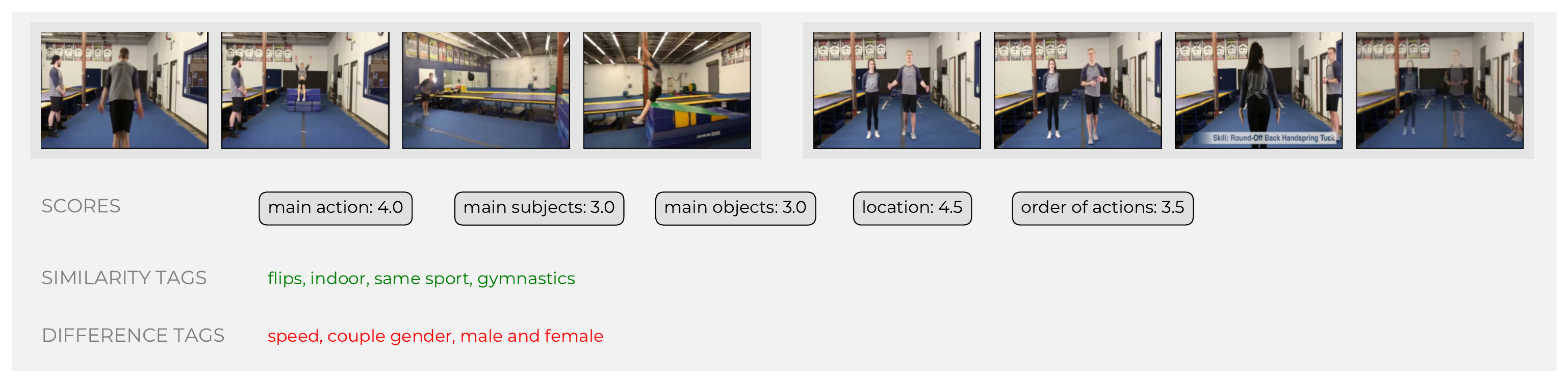}
        \caption{}
    \end{subfigure}

    \begin{subfigure}{\linewidth}
        \includegraphics[width=\linewidth]{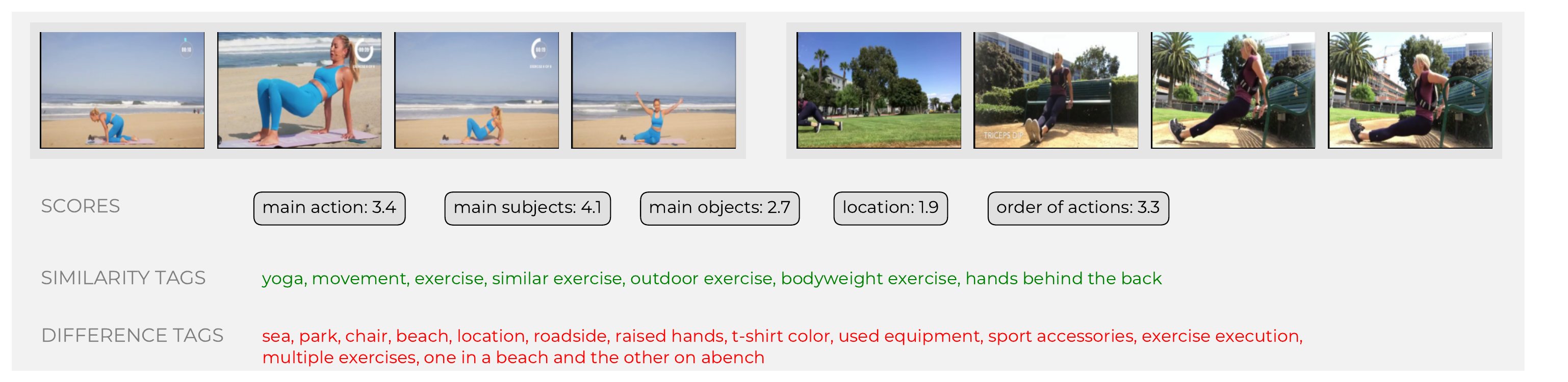}
        \caption{}
        \label{fig:qualitative_yoga_sea}
    \end{subfigure}

    \begin{subfigure}{\linewidth}
        \includegraphics[width=\linewidth]{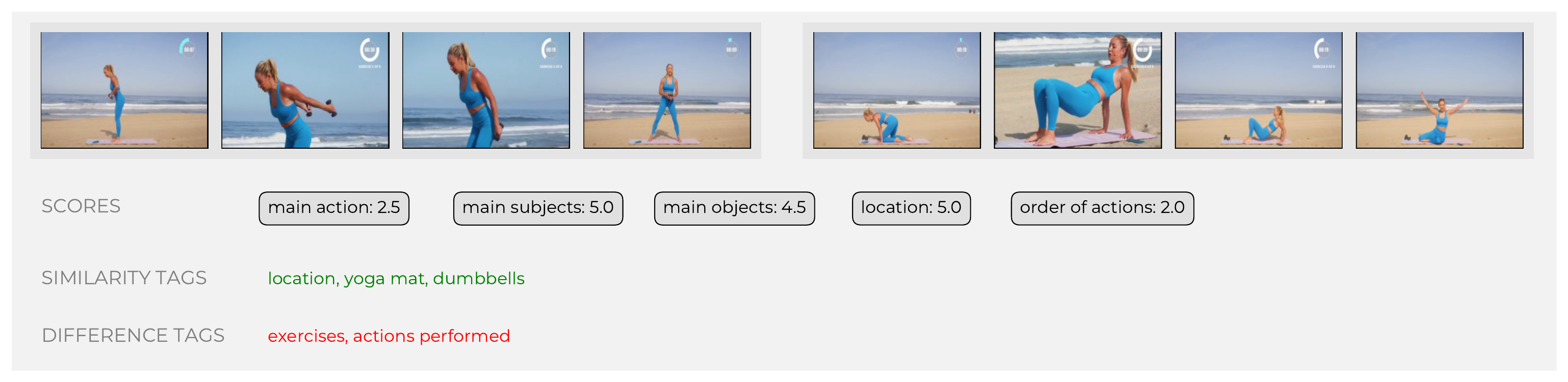}
        \caption{}
        \label{fig:qualitatives_blanket_exercises}
    \end{subfigure}
    \begin{subfigure}{\linewidth}
        \includegraphics[width=\linewidth]{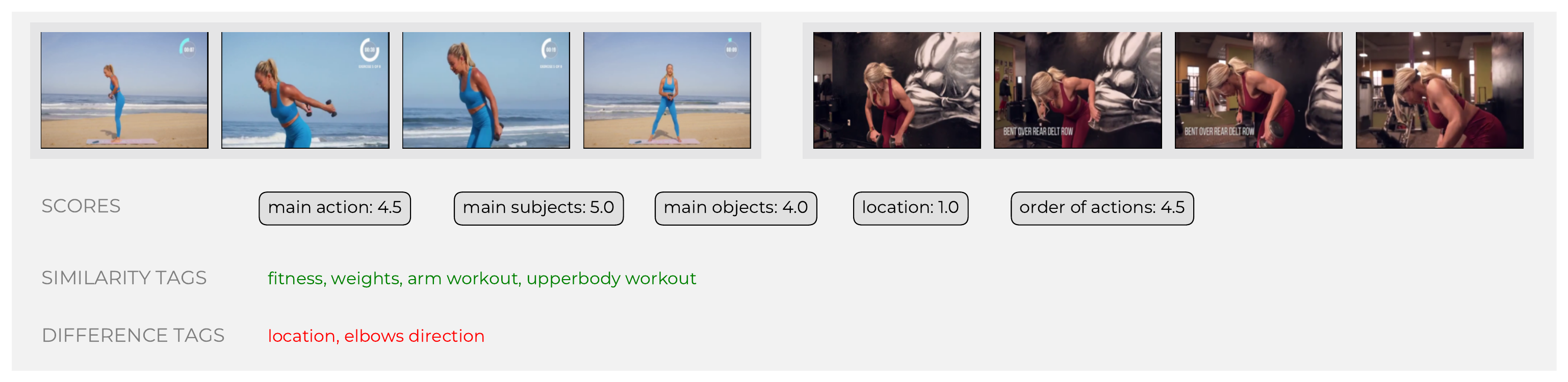}
        \caption{}
        \label{fig:qualitatives_fitness_location}
    \end{subfigure}

    \caption{\textbf{Samples from \ourdataset.}}
    \label{fig:qualitatives_part1}
\end{figure}

\begin{figure}
    \centering

    \begin{subfigure}{\linewidth}
        \includegraphics[width=\linewidth]{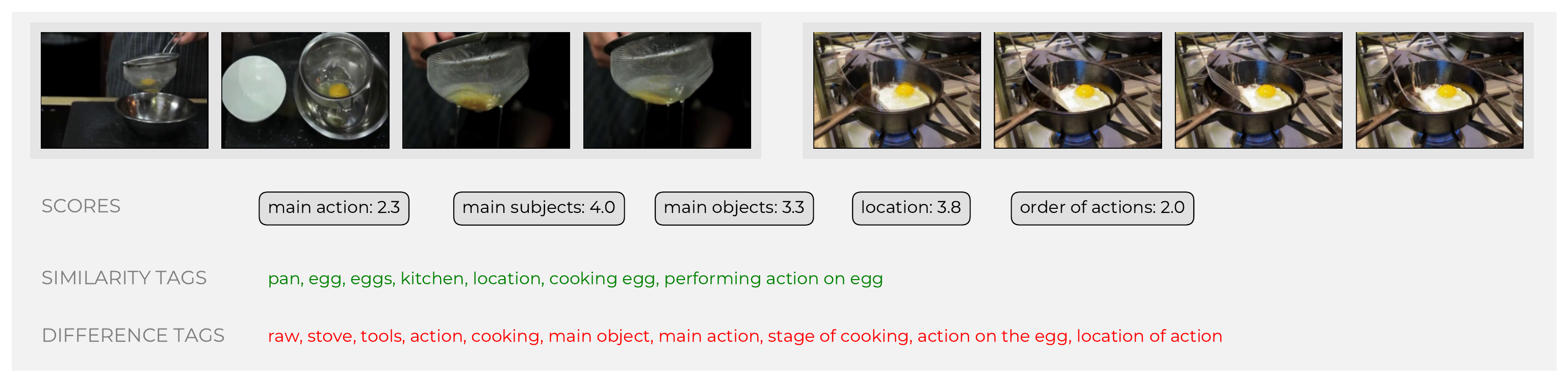}
        \caption{}
    \end{subfigure}

    \begin{subfigure}{\linewidth}
        \includegraphics[width=\linewidth]{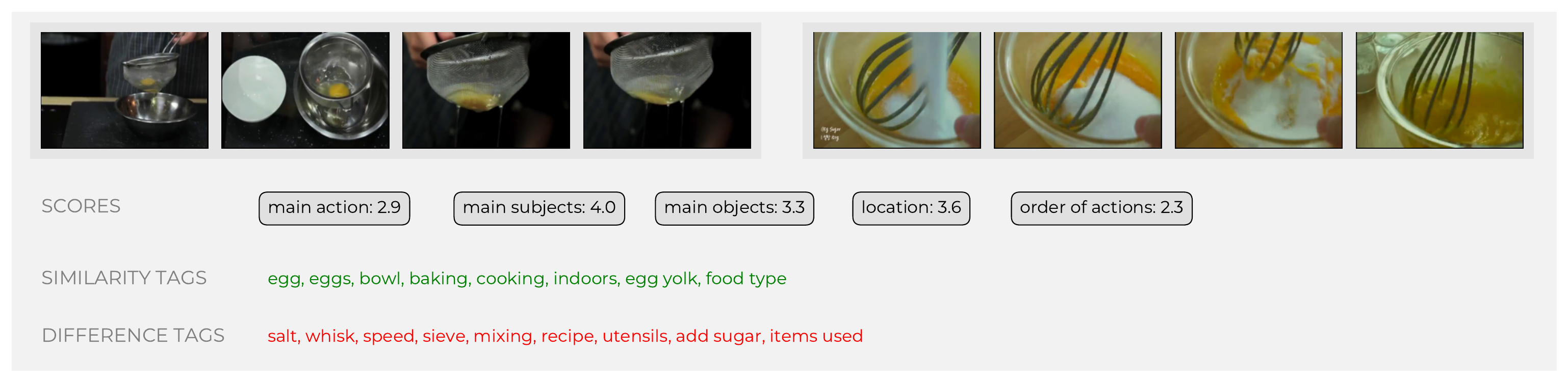}
        \caption{}
    \end{subfigure}

    \begin{subfigure}{\linewidth}
        \includegraphics[width=\linewidth]{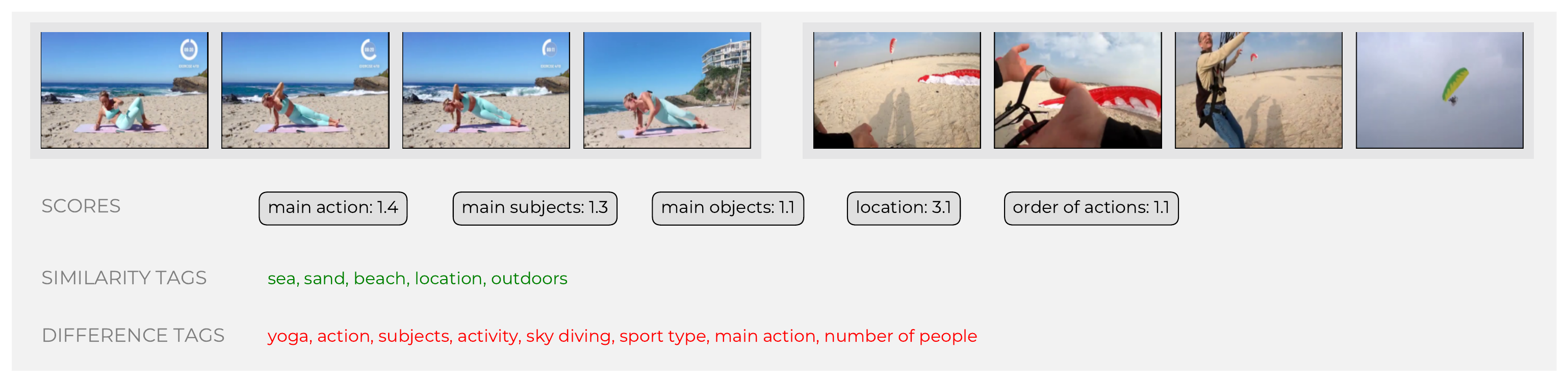}
        \caption{}
    \end{subfigure}

    \begin{subfigure}{\linewidth}
        \includegraphics[width=\linewidth]{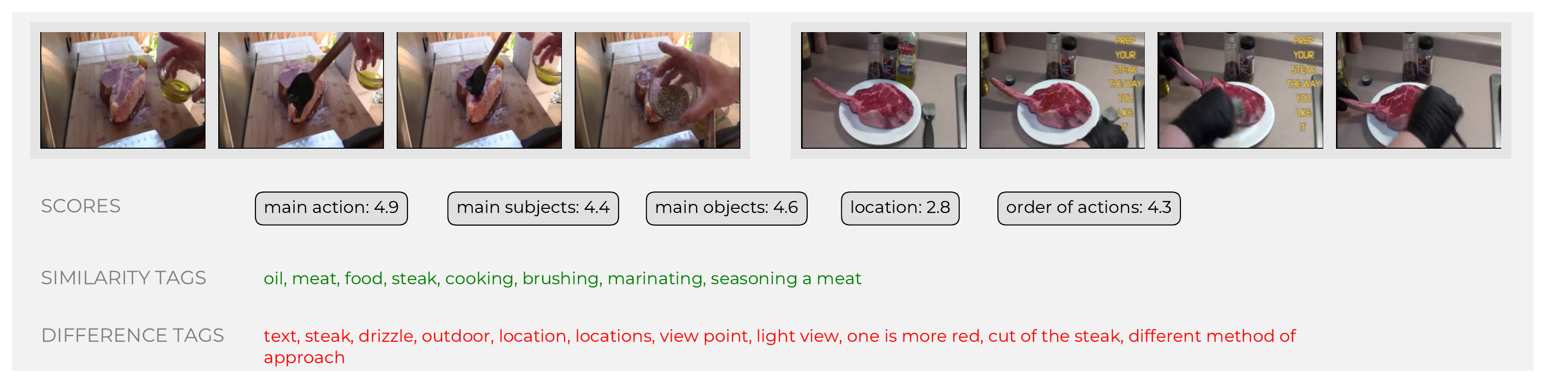}
        \caption{}
    \end{subfigure}

    \begin{subfigure}{\linewidth}
        \includegraphics[width=\linewidth]{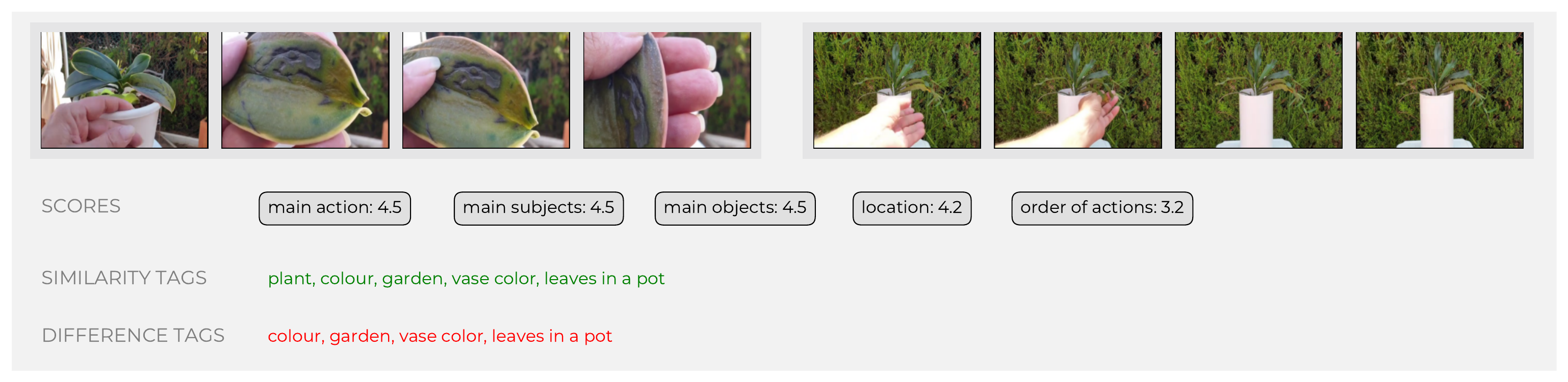}
        \caption{}
        \label{fig:qualitatives_plant_plant}
    \end{subfigure}

    \caption{\textbf{Samples from \ourdataset.}}
    \label{fig:qualitatives_part2}
\end{figure}

\begin{figure}
    \centering

    \begin{subfigure}{\linewidth}
        \includegraphics[width=\linewidth]{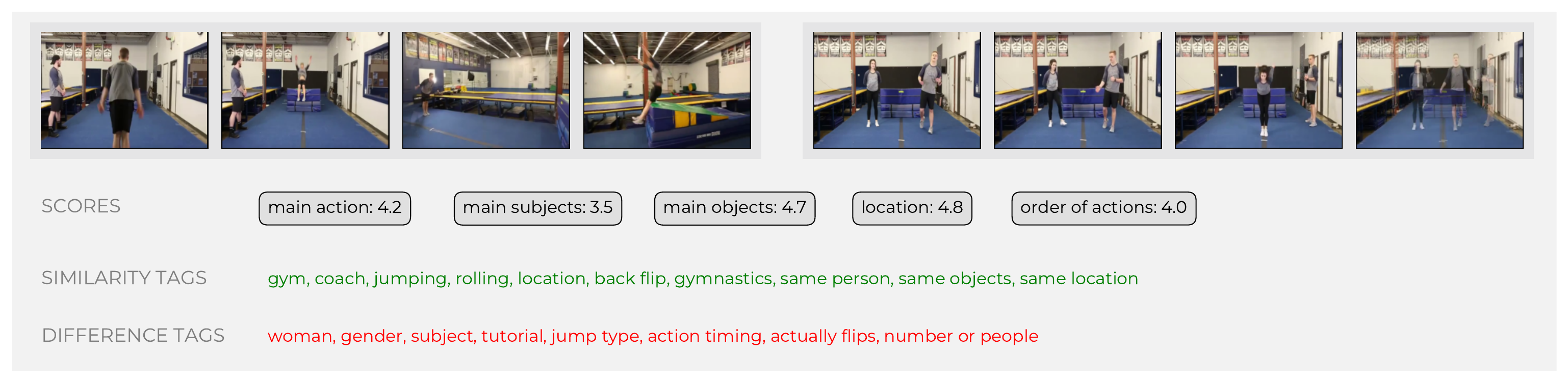}
        \caption{}
    \end{subfigure}

    \begin{subfigure}{\linewidth}
        \includegraphics[width=\linewidth]{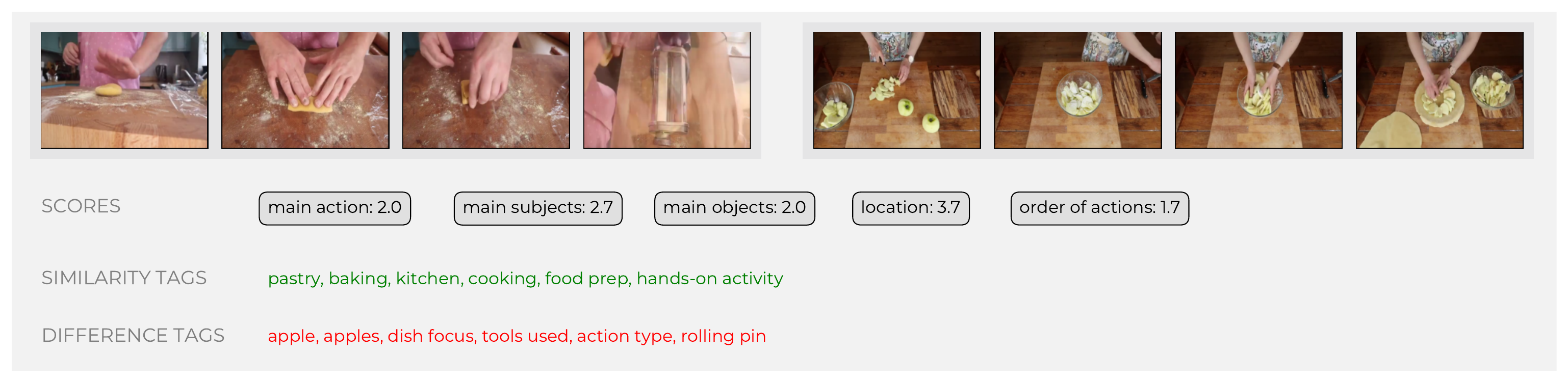}
        \caption{}
    \end{subfigure}

    \begin{subfigure}{\linewidth}
        \includegraphics[width=\linewidth]{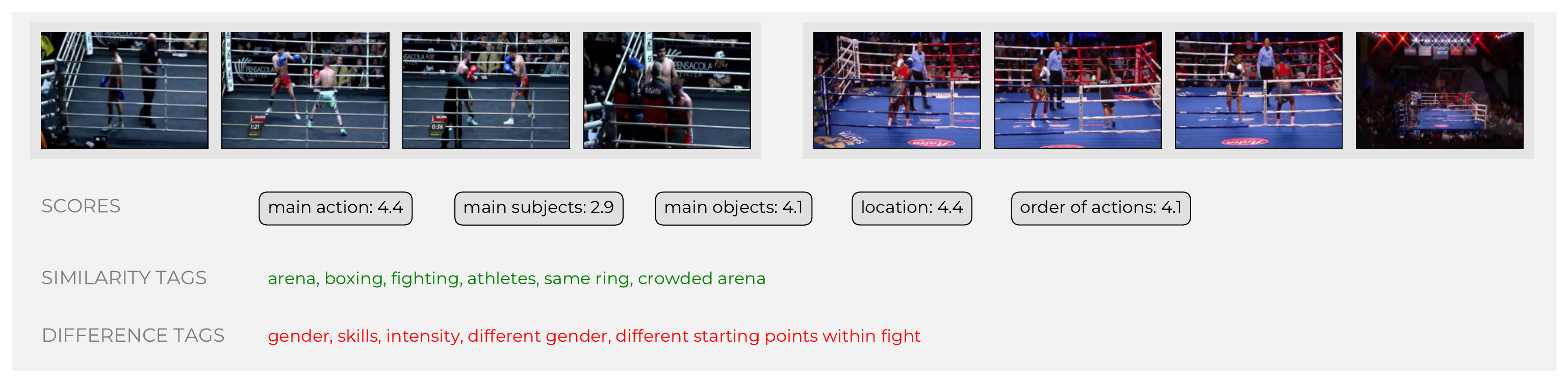}
        \caption{}
    \end{subfigure}

    \begin{subfigure}{\linewidth}
        \includegraphics[width=\linewidth]{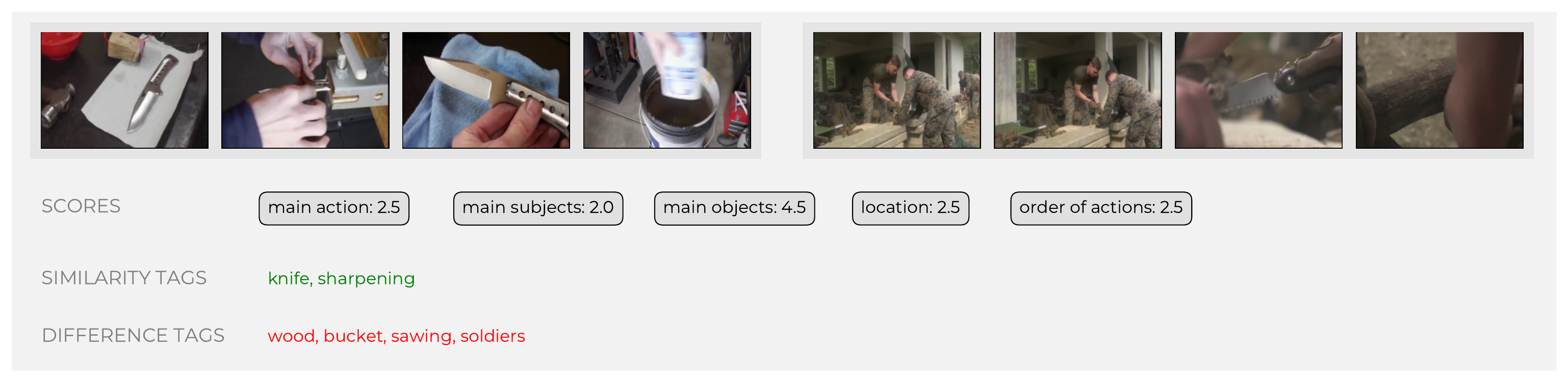}
        \caption{}
    \end{subfigure}

    \begin{subfigure}{\linewidth}
        \includegraphics[width=\linewidth]{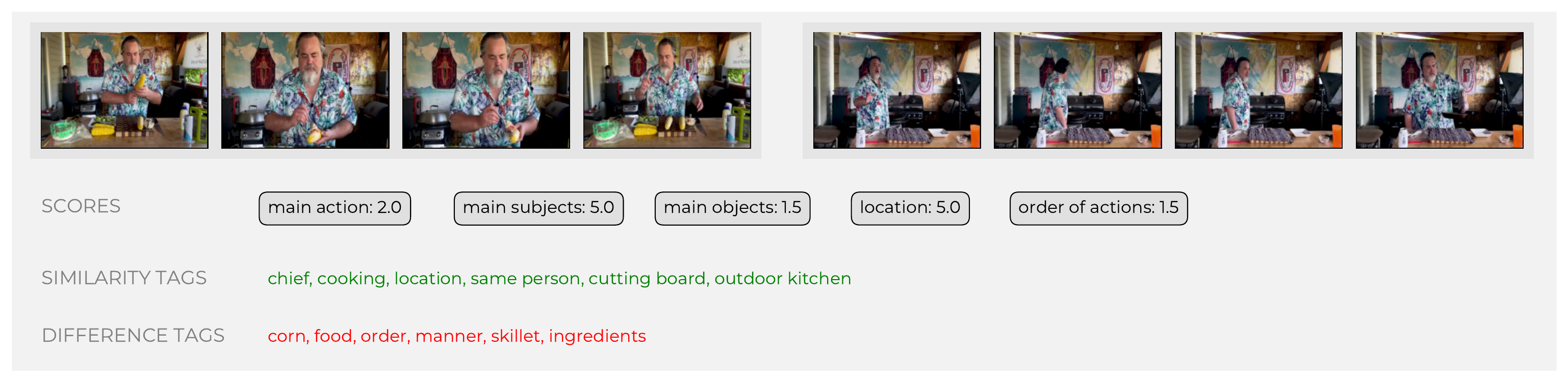}
        \caption{}
        \label{fig:qualitatives_chief_corn}
    \end{subfigure}

    \caption{\textbf{Samples from \ourdataset.}}
    \label{fig:qualitatives_part3}
\end{figure}

\begin{figure}
    \centering

    \begin{subfigure}{\linewidth}
        \includegraphics[width=\linewidth]{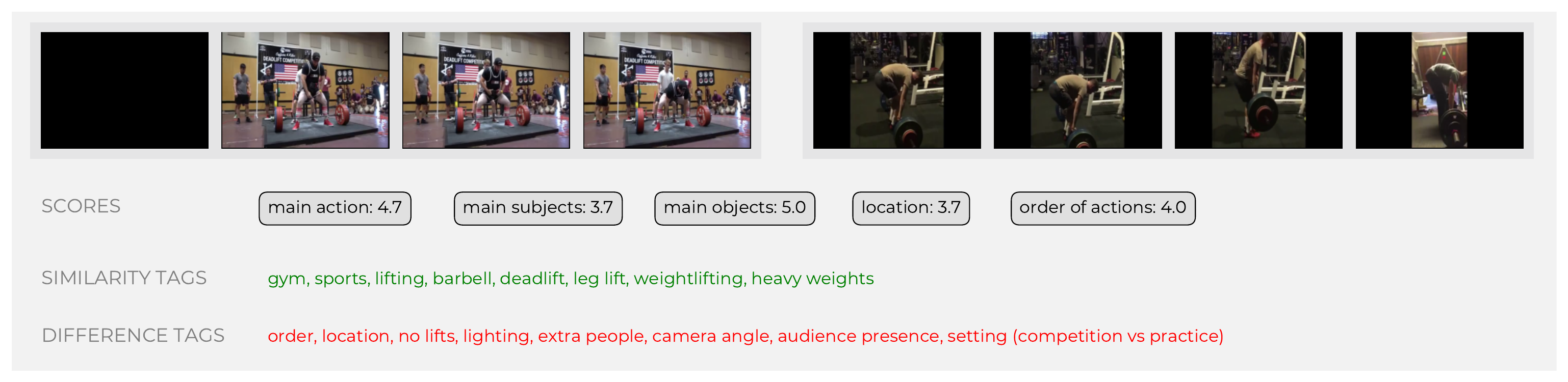}
        \caption{}
    \end{subfigure}

    \begin{subfigure}{\linewidth}
        \includegraphics[width=\linewidth]{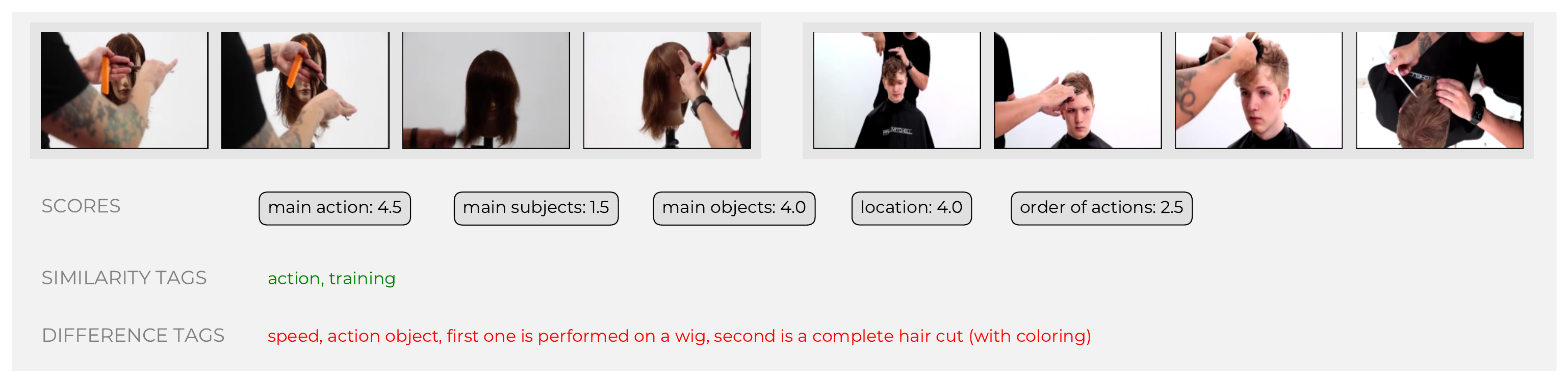}
        \caption{}
    \end{subfigure}

    \begin{subfigure}{\linewidth}
        \includegraphics[width=\linewidth]{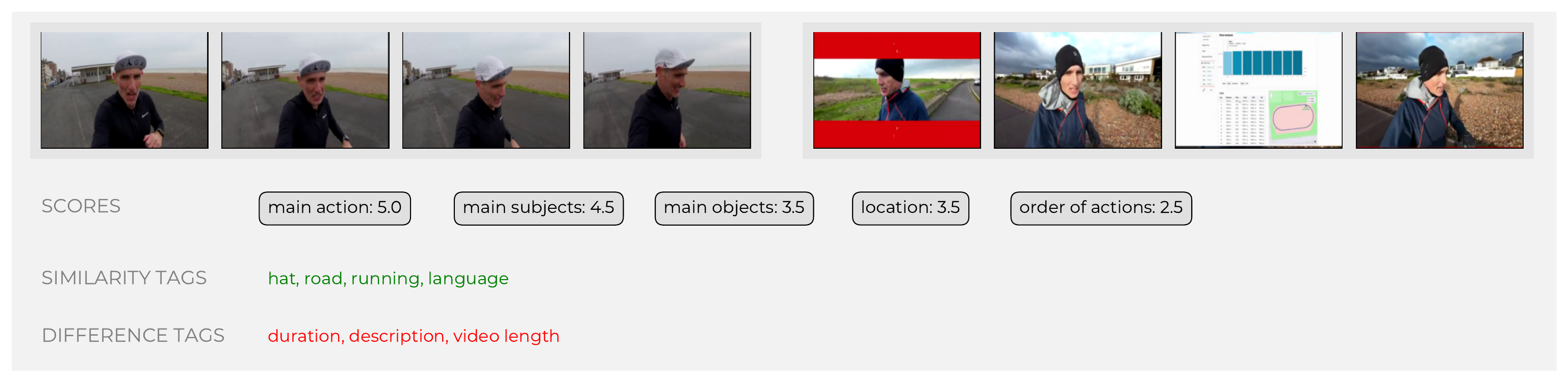}
        \caption{}
    \end{subfigure}

    \begin{subfigure}{\linewidth}
        \includegraphics[width=\linewidth]{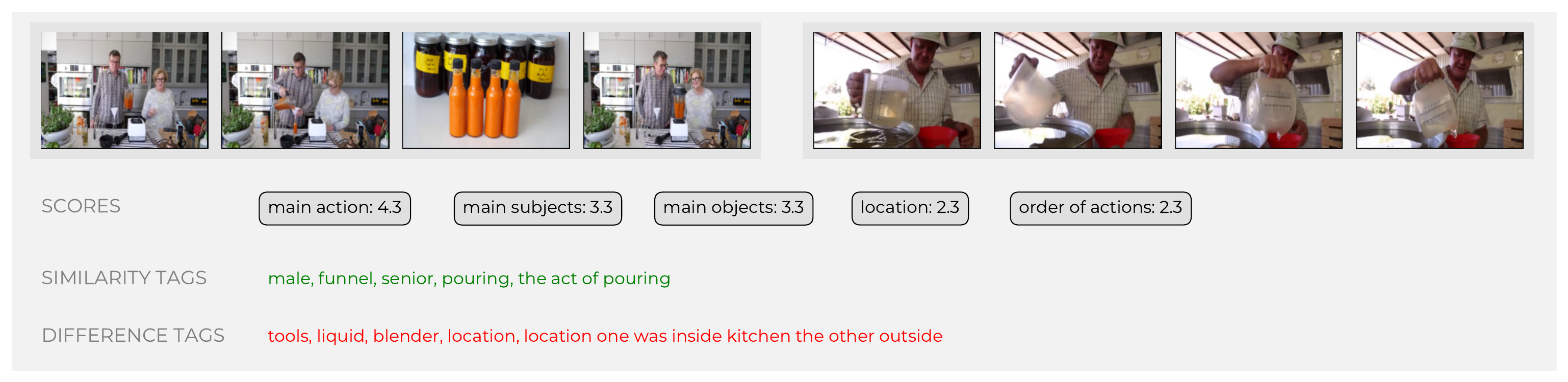}
        \caption{}
    \end{subfigure}

    \begin{subfigure}{\linewidth}
        \includegraphics[width=\linewidth]{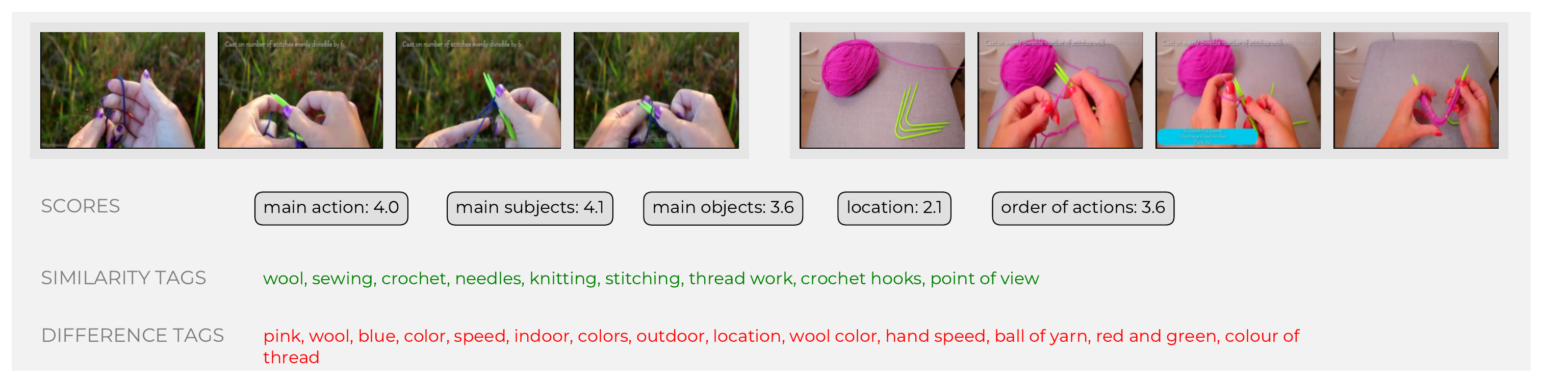}
        \caption{}
        \label{fig:qualitatives_wool_pink}
    \end{subfigure}

    \caption{\textbf{Samples from \ourdataset.}}
    \label{fig:qualitatives_part4}
\end{figure}

\begin{figure}
    \centering

    \begin{subfigure}{\linewidth}
        \includegraphics[width=\linewidth]{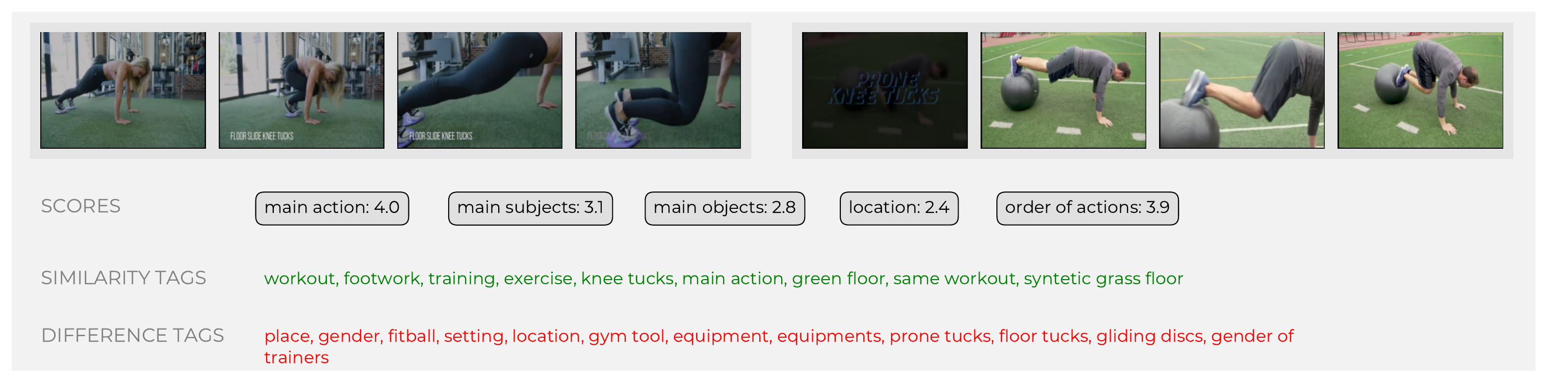}
        \caption{}
    \end{subfigure}

    \begin{subfigure}{\linewidth}
        \includegraphics[width=\linewidth]{suppfigures/1607_004.mp4-591_001.mp4_76.pdf}
        \caption{}
    \end{subfigure}

    \begin{subfigure}{\linewidth}
        \includegraphics[width=\linewidth]{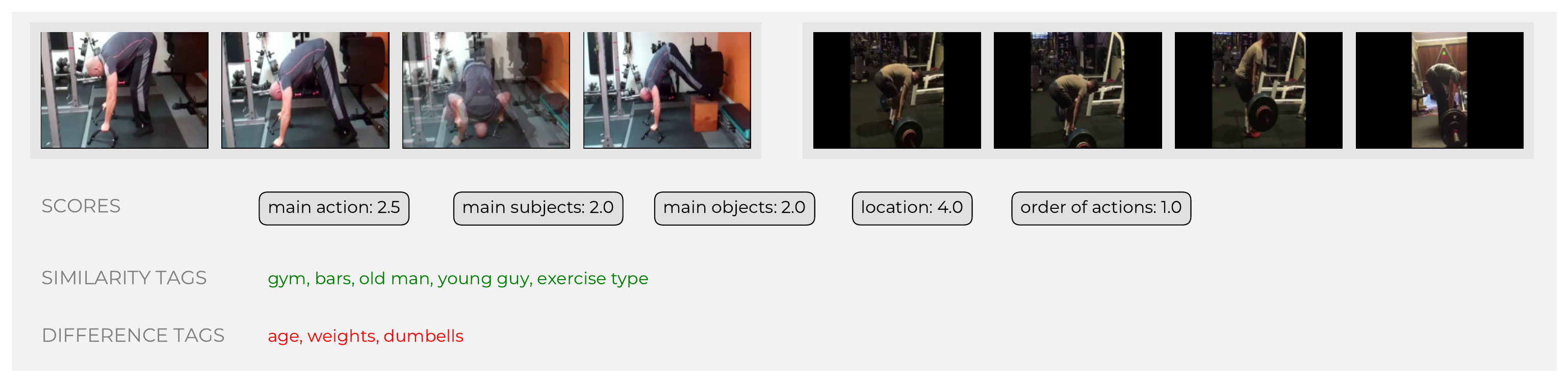}
        \caption{}
    \end{subfigure}

    \begin{subfigure}{\linewidth}
        \includegraphics[width=\linewidth]{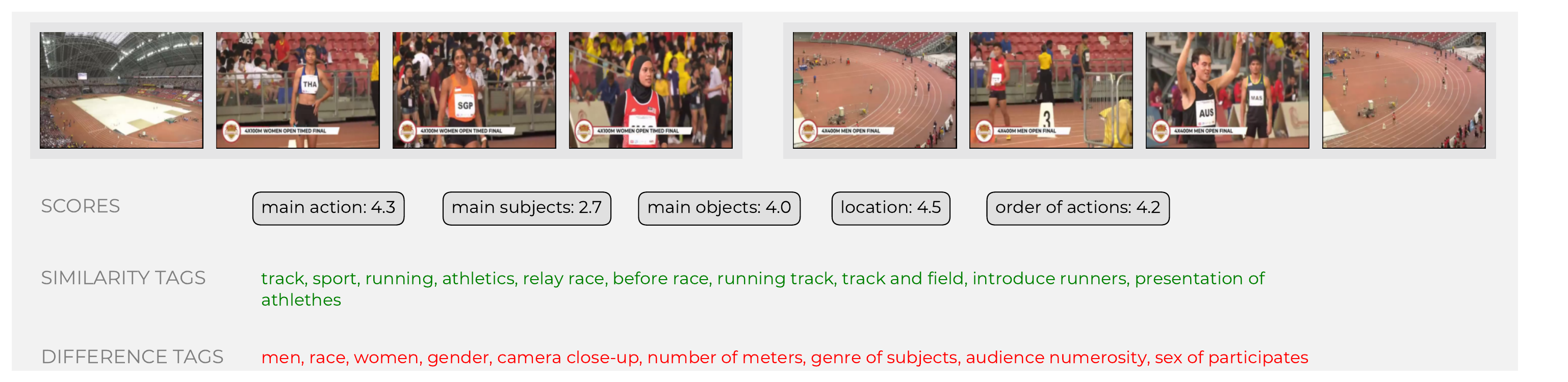}
        \caption{}
        \label{fig:qualitatives_track_men}
    \end{subfigure}

    \begin{subfigure}{\linewidth}
        \includegraphics[width=\linewidth]{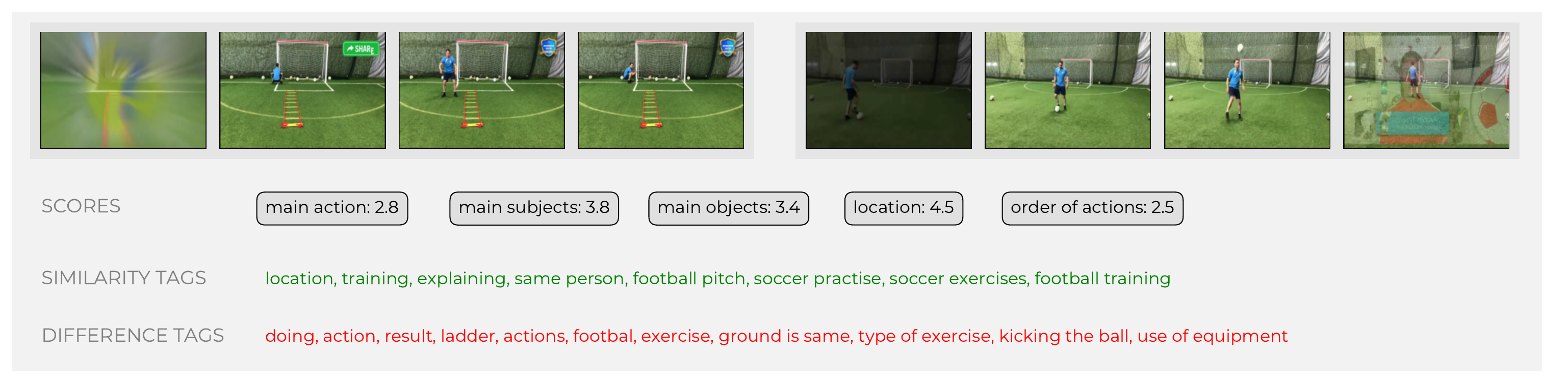}
        \caption{}
        \label{fig:qualitatives_location_doing}
    \end{subfigure}

    \caption{\textbf{Samples from \ourdataset.}}
    \label{fig:qualitatives_part5}
\end{figure}

In this section, we present qualitative examples of video pairs from \ourdataset. 
As shown in~\Cref{fig:qualitatives_part1,fig:qualitatives_part2,fig:qualitatives_part3,fig:qualitatives_part4,fig:qualitatives_part5}, each example includes two videos with four representative frames per video, together with concept scores and annotated similarity and difference tags.
\ourdataset contains similar video pairs that differ in specific aspects such as \textit{location}.
For example, a young blonde woman performing the same workout exercise at the seaside and in a park (see~\ref{fig:qualitative_yoga_sea}), 
or in an indoor gym setting (see~\ref{fig:qualitatives_fitness_location}).
Another example features a person crocheting outdoors in a park versus indoors on a sofa (see~\ref{fig:qualitatives_wool_pink}).
The dataset also includes video pairs with similar \textit{location} and \textit{main subjects}, where the \textit{main action} differs, \eg,~\ref{fig:qualitatives_blanket_exercises},~\ref{fig:qualitatives_chief_corn}, and~\ref{fig:qualitatives_location_doing} where only the action performed changes. 
Other examples show low similarity in the \textit{order of actions}, even when the overall activity or context is related. For instance, both videos may show a person interacting with plant leaves, but in one case a leaf is deliberately detached (see~\ref{fig:qualitatives_plant_plant}).
Finally, some video pairs are nearly identical in setting and action but differ in the \textit{main subjects}, such as a male versus female athlete competing in a track and field event (see~\ref{fig:qualitatives_track_men}).

Additionally, this Appendix includes qualitative examples used for the retrieval evaluation discussed in Sec.~\ref{sec:cbvr}. 
These examples are provided in video format to better demonstrate the full temporal dynamics and allow simultaneous comparison of entire videos.

\section{Broader impacts}\label{sec:broader_impacts}
The introduction of \ourtaskshort and \ourdataset presents significant potential for advancing the interpretability and alignment of models with human judgment in video understanding. 
By facilitating more reliable and human-aligned evaluation of video models, our research contributes to the development of responsible AI technologies that serve diverse communities and societal needs.
We do not foresee any immediate negative societal impacts resulting from our work.
However, there is a risk that \ourdataset may not adequately reflect the video understanding capabilities of models in specific tasks or scenarios. 
We encourage users to carefully consider the intended use cases and limitations of their models when utilizing \ourdataset for evaluation, and to supplement it with additional benchmarks or domain-relevant assessments where appropriate.

\section{Institutional review board approvals}\label{sec:review_board}
The institutional review board (IRB) reviewed and approved this study, concluding that \textit{"the research in question does not involve risks to the psycho-physical well-being of the subjects involved that could possibly even limit their right to confidentiality, information and autonomy in decision-making."}. The official document is available on request.

\section{Assets}\label{sec:assets}
Table~\ref{tab:assets} lists URLs and licenses for all the assets used in the paper. 

\begin{table}[h!]
    \centering
    \caption{\textbf{List of URLs and licenses for all assets used.}}
    \label{tab:assets}
    {\scriptsize
    \begin{tabular}{lll}
        Name & URL & License \\
        \midrule
        mPLUG-Owl3-7B & \url{https://huggingface.co/mPLUG/mPLUG-Owl3-7B-240728} &  Apache 2.0 \\ 
LLaVA-OV-7B & \url{https://huggingface.co/lmms-lab/llava-onevision-qwen2-7b-ov} & Apache 2.0 \\
LLaVA-OV-0.5B & \url{https://huggingface.co/lmms-lab/llava-onevision-qwen2-0.5b-ov} & Apache 2.0 \\
LLaVA-Video-7B & \url{https://huggingface.co/lmms-lab/LLaVA-Video-7B-Qwen2} & Apache 2.0 \\
Qwen2.5-VL-7B & \url{https://huggingface.co/Qwen/Qwen2.5-VL-7B-Instruct} & Apache 2.0 \\
Qwen2.5-VL-3B & \url{https://huggingface.co/Qwen/Qwen2.5-VL-3B-Instruct} & Apache 2.0 \\
InternVL2.5-4B & \url{https://huggingface.co/OpenGVLab/InternVL2_5-4B} & MIT \\
InternVL2.5-8B & \url{https://huggingface.co/OpenGVLab/InternVL2_5-8B} & MIT \\
InternVL3-8B & \url{https://huggingface.co/OpenGVLab/InternVL3-8B} & MIT \\
Gemini-2.0-Flash & \url{https://cloud.google.com/vertex-ai/generative-ai/docs/models/gemini/2-0-flash} & Apache 2.0 \\
VideoMAE & \url{https://huggingface.co/MCG-NJU/videomae-large} & CC BY-NC 4.0 \\
DINOv2 & \url{https://huggingface.co/facebook/dinov2-large} & Apache 2.0  \\
CLIP & \url{https://huggingface.co/openai/clip-vit-large-patch14} & MIT \\
InternVideo & \url{https://huggingface.co/OpenGVLab/InternVideo1.0} & Apache 2 \\
FineVideo & \url{https://huggingface.co/datasets/HuggingFaceFV/finevideo} & CC-BY 4.0 \\
\end{tabular} 
}
\end{table}

\clearpage
\end{document}